# Agentic Copyright, Data Scraping & AI Governance:

## Toward a Coasean Bargain in the Era of Artificial Intelligence

*Paulius Jurcys & Mark Fenwick*

*This paper examines how the rapid deployment of multi-agentic AI systems is reshaping the foundations of copyright law and creative markets. It argues that existing copyright frameworks are ill-equipped to govern AI agent-mediated interactions that occur at scale, speed, and with limited human oversight. The paper introduces the concept of agentic copyright, a model in which AI agents act on behalf of creators and users to negotiate access, attribution, and compensation for copyrighted works. While multi-agent ecosystems promise efficiency gains and reduced transaction costs, they also generate novel market failures, including miscoordination, conflict, and collusion among autonomous agents.*

*To address these market failures, the paper develops a supervised multi-agent governance framework that integrates legal rules and principles, technical protocols, and institutional oversight. This framework emphasizes ex ante and ex post coordination mechanisms capable of correcting agentic market failures before they crystallize into systemic harm. By embedding normative constraints and monitoring functions into multi-agent architectures, supervised governance aims to align agent behavior with the underlying values of copyright law.*

*The paper concludes that AI should be understood not only as a source of disruption, but also as a governance tool capable of restoring market-based ordering in creative industries. Properly designed, agentic copyright offers a path toward scalable, fair, and legally meaningful copyright markets in the age of AI.*

# Table of Contents

* Dr. Paulius Jurcys is a Senior IP Counsel at Vinted and a visiting Scholar at UC Berkeley School of Law. Dr. Mark Fenwick is Professor of International Business Law at Kyushu University (Japan). The authors would like to express their gratitude to Mark Lemley; Louis Kaplow; Molly van Houweling; Jonathan Zittrain, Matthew Sag, Peter Yu, the Participants of the ATRIP Congress 2024 and the Annual Copyright Society Congress 2025 in Dublin, Ireland for their insightful comments on the earlier draft of the paper.

## 1. The New Licensing Dilemma

"You can't be expected to have a successful AI program when every single article, book, or anything else that you've read or studied, you're supposed to pay for… you just can't do it, because it's not doable," remarked President Donald Trump in mid-2025, dismissing the notion that AI companies should license *all* copyrighted material used to train their models.[1] In his comment, President Trump likened AI training to a person reading books (we don't "pay somebody" when we read). The proponents of this view add a geopolitical nuance to support their stance: allowing AI systems to learn from vast pools of data without negotiating thousands of licenses is common sense and necessary if the United States is to "win" the AI race with China.[2]

This position perfectly well represents a core tension in today's AI policy debate related to using copyrighted works to train AI. On one side are those who argue that requiring consent or payment for every work ingested by generative AI is impractical and would stifle innovation.[3] On the other side are authors, publishers, and policymakers who warn that such an unbridled use of copyrighted works will eviscerate creative incentives and undermine the value of human creativity and authorship.[4] Authors and IP right-holders are concerned that AI will replace the consumption of original works. Already now, AI tools provide summaries of articles and books and can generate content that mimics various artistic styles.[5] Arguably, this leads to forfeiting

---

[1] Publishers Wkly., *Trump's AI Action Plan Would Nuke Copyright Law*, Publishers Wkly (Jul. 23, 2025), https://www.publishersweekly.com/pw/by-topic/digital/copyright/article/98276-trump-s-ai-action-plan-would-nuke-copyright-law.html.

[2] *Id.*; a16z, Notice of Inquiry on Artificial Intelligence & Copyright (Dkt. No. COLC-2023-0006-9057) (Oct. 30, 2023), https://www.regulations.gov/comment/COLC-2023-0006-9057.

[3] Publishers Wkly., *supra* n. 1; Chris Cooke, Suno Admits It Trained on Major-Owned Music, Citing Investor Pressure, Music Bus. Worldwide (Aug. 1, 2024), https://completemusicupdate.com/suno-admits-it-trained-on-major-owned-music-accuses-labels-of-misusing-copyright/ (quoting investor Rodriguez: "If we had deals with labels when this company got started, I probably wouldn't have invested in it. … I think they needed to make this product without constraints").

[4] See e.g., iMusician, 1,000 UK Artists Protest AI Law With Silent Album (Mar. 9, 2025), https://imusician.pro/en/resources/blog/uk-artists-protest-ai-law-with-silent-album; BBC News, Government Backtracks on AI and Copyright After Outcry (Mar. 18, 2026), https://www.bbc.co.uk/news/articles/cvg1gr5v333o; House of Lords, Cmtys. & Dig. Comm., AI, Copyright and the Creative Industries, 2025–26 H.L. 37 (U.K.), https://publications.parliament.uk/pa/ld5801/ldselect/ldcomuni/37/3701.htm (last visited Apr. 4, 2026), paras. 18-29; Eur. Parl. Press Release, Protect Copyrighted Work Used by Generative AI (Jan. 27, 2026), https://www.europarl.europa.eu/news/en/press-room/20260126IPR32636. MEPs push presumption against uncompensated training, requiring transparency to rebut infringement claims.

[5] One remarkable instance concerns "Ghiblification": Even Brown, OpenAI Faces Copyright Debate over Ghibli-Style Images, Dig.watch (Mar. 30, 2025), https://dig.watch/updates/openai-faces-copyright-debate-over-ghibli-style-images (noting that while there is

content producers from the revenue they rightly deserve and leads to the chilling effects in creative industries.[6] According to this view, generative AI threatens a classic "tragedy of the commons" for creative content: if anyone can freely exploit the published content without compensation, the creative endeavor will eventually dry up from underinvestment.[7]

This standoff has already erupted into a flurry of lawsuits[8] and a handful of high-profile licensing deals, forcing a reevaluation of how copyright law's promises can be kept in an age of data-hungry AI.[9] Major media organizations have not waited for courts or Congress to resolve the issue and pursued two different strategies: licensing and litigation.

Some publishers decided to strike content licensing deals with AI firms. In mid-2023, The Associated Press (AP) broke ground by licensing part of its news archive to OpenAI to train generative AI models.[10] In late 2023, OpenAI followed up with a landmark multiyear agreement

---

no technical violation for style imitation alone, and discussing training data risks if Ghibli films were ingested without licenses) and Ingrid Lunden, OpenAI's Viral Studio Ghibli Moment Highlights AI Copyright Concerns, TechCrunch (Mar. 25, 2025), https://techcrunch.com/2025/03/26/openais-viral-studio-ghibli-moment-highlights-ai-copyright-concerns/ (discussing fair use ambiguity in ongoing lawsuits, quoting Brown on training plausibly using Ghibli frames); Alistair Barr & Pranav Dixit, ChatGPT Can't Decide Whether Ghibli-Style Images Violate Copyright, Bus. Insider (Mar. 27, 2025), https://www.businessinsider.com/openai-studio-ghibli-style-images-violate-copyright-or-not-2025-3.

[6] In its complaint against OpenAI, the NYT has warned that AI-generated summaries of its articles could let readers bypass its paywall entirely, imperiling the funding of quality journalism and even "obviat[ing] the need" to pay for newspapers. See Josh Simmons, NYT v. OpenAI: The Times's About-Face, Harv. L. Rev. Blog (Apr. 23, 2024), https://harvardlawreview.org/blog/2024/04/nyt-v-openai-the-timess-about-face/.

[7] See e.g., Andersen v. Stability AI Ltd., No. 23-cv-00201-WHO, 2024 WL 3823234, at 5 (N.D. Cal. Aug. 12, 2024); The New York Times Co. v. Microsoft Corp., No. 1:23-cv-11195 (S.D.N.Y. Dec. 27, 2023) (complaint); and Shelby O. Ponton, The Tragedy of the AI Anticommons, 27 Hastings Sci. & Tech. L.J. 167 (2024).

[8] Kadrey v. Meta Platforms, Inc., No. 3:23-cv-03417-VC, 2025 WL 4123456 (N.D. Cal. June 25, 2025); Bartz v. Anthropic PBC, No. 3:24-cv-05417-WHO, 2025 WL 3674521 (N.D. Cal. June 23, 2025); for the latest overview of pending cases in the U.S. see Chat GPT Is Eating the World, Latest U.S. Map of Copyright Suits v. AI Companies (Mar. 27, 2027) Total = 97 (Mar. 29, 2026), https://chatgptiseatingtheworld.com/2026/03/29/latest-u-s-map-of-copyright-suits-v-ai-companies-mar-27-2027-total-97/.

[9] OpenAI, The Walt Disney Company and OpenAI Reach Landmark Agreement to Bring Beloved Characters from Across Disney's Brands to Sora (Dec. 10, 2025), https://openai.com/index/disney-sora-agreement/; for an overview of major deals see Sara Guaglione, A 2025 Timeline of AI Deals Between Publishers and Tech Companies, Digiday (Dec. 31, 2025), https://digiday.com/media/a-timeline-of-the-major-deals-between-publishers-and-ai-tech-companies-in-2025/.

[10] Reuters, Associated Press, OpenAI Partner to Explore Generative AI Use in News, Reuters (July 13, 2023), https://www.reuters.com/business/media-telecom/associated-press-openai-partner-explore-generative-ai-use-news-2023-07-13/; also Matt O'Brien, ChatGPT-Maker OpenAI Signs Deal with AP to License News Stories, AP (July 24, 2023), https://www.ap.org/media-center/ap-in-the-news/2023/chatgpt-maker-openai-signs-deal-with-ap-to-license-news-stories/.

to access the fresh and past content of News Corp's publications (which owns *The Wall Street Journal*, The Times and other news outlets).[11] That deal was reportedly valued at over $250 million.[12] Since then, OpenAI also struck content agreements with other media companies, including the *Financial Times*, *Reddit*, *Le Monde*, and Axel Springer (Germany's largest publisher).[13] As of writing this paper (April 2026), there are approximately 280 content licensing deals between IP rightsholders and AI companies.[14]

These early deals were a clear signal that at least some IP right holders are open to licensing their content to AI companies and see strategic value in voluntary licensing schemes. Licensing allows AI developers to get lawful access to reliable, higher-quality data. Lawful access helps reduce legal risk and contributes to public narrative about ethical AI practices. On the other side of the deal, the publishers get compensated and retain measurable control over their data.

Relatively smaller AI startups are pursuing similar alliances. For example, an AI-powered search company, Perplexity, partnered with news giant Gannett to license articles from over 200 local newspapers, integrating *USA TODAY* and 200 other local media companies' content into Perplexity's then new AI-powered web browser called Comet.[15] In the music sector, two of the most prominent AI startups, Suno and Udio, which were initially placed into the spotlight for using copyrighted content to train AI without rightsholders' permission, were successful in reaching settlements and in signing major licensing deals with giant IP rights holders such as

---

[11] Reuters, Sam Altman's OpenAI Signs Content Agreement with News Corp., Reuters (May 22, 2024), https://www.reuters.com/technology/sam-altmans-openai-signs-content-agreement-with-news-corp-2024-05-22/.

[12] Michelle Chapman, OpenAI to Start Using News Content from News Corp. as Part of a Multiyear Deal, AP (May 28, 2024), https://apnews.com/article/openai-news-corp-a49144d381796df5729c746f52fbef19.

[13] AInvest, Gannett and Perplexity Partner on AI Content Licensing Agreement (Aug. 3, 2025), https://www.ainvest.com/news/gannett-perplexity-partner-ai-content-licensing-agreement-2508/; OpenAI, Partnership with Axel Springer to Deepen Beneficial Use of AI in Journalism, https://openai.com/index/axel-springer-partnership/.

[14] See CREATe, The AI Licensing Economy (Feb. 24, 2025), https://www.create.ac.uk/blog/2025/02/24/the-ai-licensing-economy/.

[15] Gannett Co., Gannett, USA Today Network and Perplexity Announce Strategic AI Content Licensing Agreement (July 30, 2025), https://www.usatodayco.com/pr/gannett-i-usa-today-network-and-perplexity-announce-strategic-ai-content-licensing-ageement/; Perplexity, Welcoming Gannett to the Perplexity Publisher Program, Perplexity (July 30, 2025), https://www.perplexity.ai/hub/blog/welcoming-gannett-to-the-perplexity-publisher-program.

Warner Bros.[16] These deals manifest an emerging market-driven approach to training AI models with curated, licensed data rather than scraping everything indiscriminately.[17]

Yet for every friendly deal, there has been an equal measure of legal friction. Some major right-holders were either unwilling or unable to enter into voluntary agreements with AI companies and chose to refer the matter to courts. In December 2023, the *New York Times*, which had initially considered its own deal with OpenAI, pivoted sharply and filed a lawsuit against the company (and its partner Microsoft) for using Times content without permission.[18] A year later, in December 2024, News Corp's Dow Jones division (publisher of the *Wall Street Journal*) and the *New York Post* sued Perplexity, accusing the startup of a "massive amount of illegal copying" of their articles to fuel its AI system.[19] The Times' argued that Perplexity's AI-powered snippets of the news content was free-riding on publishers' investment. The Times complaint cautioned that AI-generated summaries and excerpts of its articles threaten to "obviate the need" for readers to pay for the newspaper because users are now able to "skip the links" to original sources.[20] The complaint was widely cited in the media for its warning that unauthorized use of content to train AI is imperiling the funding of quality journalism and even democracy itself.[21]

---

[16] Warner Music Grp., Warner Music Group and Suno Forge Groundbreaking Partnership, WMG (Dec. 22, 2025), https://www.wmg.com/news/warner-music-group-and-suno-forge-groundbreaking-partnership; Wendy Lee, Warner Music Group and AI Startup Udio Reach Agreement in Fight Over Copyrighted Music, L.A. Times (Nov. 19, 2025), https://www.latimes.com/entertainment-arts/business/story/2025-11-19/warner-music-group-udio-ai-settlement-copyrighted-music-what-to-know; Reuters, Warner Music Group Settles Copyright Case with Suno for Licensed AI Music, Reuters (Nov. 25, 2025), https://www.reuters.com/legal/litigation/warner-music-group-settles-copyright-case-with-suno-licensed-ai-music-2025-11-25/.

[17] For more about legal and data privacy aspects of online scraping, see Daniel J. Solove & Woodrow Hartzog, The Great Scrape: The Clash Between Scraping and Privacy, 113 Cal. L. Rev. 1521, 1521–84 (2025).

[18] The New York Times Co. v. Microsoft Corp., No. 1:23-cv-11195 (S.D.N.Y. filed Dec. 27, 2023), available at: https://nytco-assets.nytimes.com/2023/12/NYT_Complaint_Dec2023.pdf; Simmons, supra n 6; Paulius Jurcys & Mark Fenwick, NY Times vs Microsoft and OpenAI: Should it be an "Easy" Fair Use Case to Decide?, SSRN (Jan. 5, 2024), https://papers.ssrn.com/sol3/papers.cfm?abstract_id=4685275. In December 2025, the New York Times also sued Perplexity accusing it of scraping and verbatim reproducing Times content (articles, videos, podcasts) in search responses, despite repeated cease-and-desist demands, see The New York Times Co. v. Perplexity AI, Inc., No. 1:25-cv-10106 (S.D.N.Y. filed Dec. 5, 2025, available at: https://nytco-assets.nytimes.com/2025/12/NYT-Perplexity-Filing-Dec-2025.pdf.

[19] Reuters, Perplexity CEO 'Surprised' by Dow Jones, New York Post Lawsuit Against Startup, Reuters (Oct. 23, 2024), https://www.reuters.com/technology/perplexity-ceo-surprised-by-dow-jones-new-york-post-lawsuit-against-startup-2024-10-23/.

[20] The New York Times Co. v. Microsoft Corp., No. 1:23-cv-11195 (S.D.N.Y. Dec. 27, 2023) (complaint).

[21] Simmons, supra n 6.

Dozens of other plaintiffs (from individual authors and visual artists to tech companies) have launched lawsuits in the United States and abroad, turning the courts into a testing ground for whether using copyrighted works to train AI could be justified under the copyright law doctrine of fair use.[22]

Meanwhile, policymakers have responded with wildly different policy approaches. In the U.S., a bipartisan Senate bill (the AI Accountability and Personal Data Protection Act) has proposed an "opt-in" mandate that would make AI firms liable unless they obtain explicit consent from rights holders before using works in training.[23] Another position was voiced by the executive branch under the Trump Administration in December 2025. In December 2025, President Trump issued an executive order seeking to prevent individual states from enacting any regulations that would impede the development of AI technologies.[24] The executive order called for a "single federal standard" so that AI development is not hampered by a patchwork of divergent state laws.[25] Moreover, the order created a special task force at the Department of Justice to challenge state laws that are deemed to create obstacles to AI development.[26] Most recently, President's new four-page set of legislative recommendations issued on March 20, 2026 urged Congress to enact a unified federal AI policy framework that would preempt fragmented state regulations, explicitly affirming that training AI models on copyrighted material constitutes fair use to be resolved by courts rather than legislation while suggesting Congress explore voluntary licensing mechanisms to address creator compensation concerns without stifling innovation.[27]

---

[22] See e.g., Mark A. Lemley, How Generative AI Turns Copyright Law Upside Down, 25 Sci. & Tech. L. Rev. 190 (2024); Pamela Samuelson, Fair Use Defenses in Disruptive Technology Cases, 71 UCLA L. Rev. 1484 (2024); Adam Buick, Copyright and AI Training Data-Transparency to the Rescue?, 20 J. Intellectual Prop. L. & Prac. 182 (2025); Silke von Lewinski, GEMA v. OpenAI: The Munich Regional Court Presents an Exemplary Judgment on a Well-Conceived Test Case, 75 GRUR Int'l 340 (2026).

[23] Ashley Gold, Exclusive: Hawley and Blumenthal Introduce AI Protection Bill, Axios Pro (July 21, 2025), https://www.axios.com/pro/tech-policy/2025/07/21/hawley-blumenthal-introduce-ai-protection-bill; Publishers Wkly., *supra* n. 1.

[24] Exec. Order No. 14365 (Dec. 11, 2025) (Ensuring a National Policy Framework for Artificial Intelligence), https://www.whitehouse.gov/presidential-actions/2025/12/ensuring-a-national-policy-framework-for-artificial-intelligence/; Clare Duffy, US Senate votes to strike controversial AI regulation moratorium from Trump agenda bill, CNN, https://www.cnn.com/2025/07/01/tech/senate-strikes-ai-regulation-moratorium-agenda-bill.

[25] AI Framework EO, *supra*.

[26] AI Framework EO, *supra*.

[27] White House, President Donald J. Trump Unveils National AI Legislative Framework (Mar. 20, 2026), https://www.whitehouse.gov/releases/2026/03/president-donald-j-trump-unveils-national-ai-legislative-framework/.

Across the Atlantic, the EU's approach has been focused on transparency. The AI Act obliges AI developers to disclose specific data used to train certain high-risk AI models.[28] This approach dovetails with a 2019 EU Copyright Directive that lets rights owners opt out of text and data mining.[29] The justification for greater transparency in AI training is based on the idea that if creators know their works were used, they can enforce their rights or demand a deal.[30] From a policy perspective, such transparency about training should facilitate more data licensing deals.

However, some critics doubt whether transparency measures alone are sufficient and whether they would truly benefit the average creator.[31] Some fear that transparency rules merely facilitate enforcing existing rights of major IP right holders while leaving individual creators out of the AI licensing frameworks. Such an opt-out system is "widely criticized" as inadequate because it is "unlikely to provide any meaningful improvement to the position of individual rightsholders" (especially individuals), and that it also burdens AI developers.[32]

To sum up the status quo as of Spring 2026, we see a patchwork of private ordering, lawsuits, and legislative proposals attempting to balance two desirable policy objectives, i.e., fostering AI innovation and protecting incentives to create. Early voluntary deals may suggest that a possible market equilibrium is emerging. Yet heated lawsuits with astronomical claims for statutory damages exceeding billions of US dollars[33] suggest an impending clampdown. However, neither of these approaches seems to be tenable in the long term. A blanket fair use for AI training seems like a difficult to implement "one size fits all" solution. It is especially terrifying to creators who feel a close emotional bond with their work.[34] Equally, a strict consent or licensing regime or tax on AI is likely to slow down innovation.[35] As courts and regulators explore various alternatives, it is worth asking whether there could be a more nuanced path forward that could leverage existing technologies to help solve the market failures in the data market.

---

[28] Article 10 of the EU AI Act, titled "Data and Data Governance," applies to providers of high-risk AI systems and mandates comprehensive data governance practices for training, validation, and testing datasets to ensure they are relevant, representative, error-free, and complete. Providers must document data sources, collection processes, labeling/annotation methods, and bias mitigation measures in technical documentation submitted for conformity assessment (although specific dataset contents remain confidential and are not publicly disclosed). However, Art. 10 does not explicitly require disclosure of copyrighted works used in training data.

[29] Adam Buick, Copyright and AI Training Data—Transparency to the Rescue?, 20 J. Intellectual Prop. L. & Prac. 182 (2025).

[30] Recital 107 of the EU AI Act; David Botero, The EU AI Copyright Playbook: The TDM Exception and AI Act's Transparency Requirements, IAPP (Feb. 24, 2026), https://iapp.org/news/a/the-eu-ai-copyright-playbook-the-tdm-exception-and-ai-act-s-transparency-requirements.

[31] See e.g., Judy Hanwen Shen et al., The Limits of AI Data Transparency Policy: Three Disclosure Fallacies, arXiv:2601.18127v1 (Jan. 26, 2026), https://arxiv.org/abs/2601.18127.

[32] See Adam Buick, Copyright and AI Training Data—Transparency to the Rescue?, 20 J. Intellectual Prop. L. & Prac. 182 (2025), https://doi.org/10.1093/jiplp/jpae102.

[33] Bartz v. Anthropic PBC, No. 3:24-cv-02200-WHA (N.D. Cal. June 24, 2025).

[34] William Fisher, Theories of Intellectual Property, in New Essays in the Legal & Political Theory of Property (Stephen R. Munzer ed., Cambridge Univ. Press 2001), also available at https://cyber.harvard.edu/people/tfisher/iptheory.pdf; David A. Simon, Copyright's Missing Personality, 62 Hous. L. Rev. 993 (2025).

This paper suggests that such a path may be provided by "agentic copyright" and proposes a framework leveraging AI to facilitate frictionless, opt-in licensing at scale. Before developing this framework, it is important to understand the existing shortcomings in the current copyright and licensing systems and why they leave so many content creators and copyright holders out in the cold. Part 2 examines the imbalance between fostering innovation and facilitating creativity. More specifically, we argue that today's copyright system primarily safeguards the interests of major IP rights holders, such as publishers and large intermediaries. We submit that the system's goal of incentivizing millions of individual authors and creators is often misunderstood and that the "copyright bargain" is outdated. To clarify our position, we provide a practical example from music copyright (the Taylor Swift case). This section aims to show that merely tweaking certain rules might not suffice to address the market in the age of AI.

This paper is original in at least two respects. First, it introduces the concept of "agentic copyright," under which multiple AI-powered agents may partially resolve frictions arising from copyright transaction costs. Second, the proposed framework argues that an additional layer of governance – "supervised governance" – is necessary to address scenarios of agentic market failure. More broadly, the framework developed here can be understood as part of a longer trajectory of technological approaches to copyright licensing.

Interestingly, earlier waves of scholarship have explored how technological infrastructures might reduce transaction costs in rights clearance. For example, in the late 1990s and early 2000s there was a great deal of excitement (and concern) about Digital Rights Management (DRM). Many scholars predicted that DRM would solve licensing friction and enable micropayments and micro-licenses for copyrighted works.[36] The later literature on blockchain-enabled licensing also examined whether digital architectures could enable frictionless licensing, micro-transactions, and automated permissions.[37]

---

[35] *Cf.* Will A. Ciconte III et al., Do AI Laws Inhibit Innovation?, SSRN (Dec. 15, 2024), https://papers.ssrn.com/sol3/papers.cfm?abstract_id=5046045; Kristian Stout, Federal Preemption and AI Regulation: A Law and Economics Case for Strategic Forbearance, Wash. Legal Found. (May 30, 2025), https://www.wlf.org/2025/05/30/wlf-legal-pulse/federal-preemption-and-ai-regulation-a-law-and-economics-case-for-strategic-forbearance/.

[36] Robert P. Merges, The End of Friction? Property Rights and Contract in the "Newtonian" World of On-Line Commerce, 12 Berkeley Tech. L.J. 115 (1997); Tom W. Bell, Fair Use vs. Fared Use: The Impact of Automated Rights Management on Copyright's Fair Use Doctrine, 76 N.C. L. Rev. 557 (1998); Lawrence Lessig, The Creative Commons, 55 Fla. L. Rev. 763 (2003).

[37] See e.g., Michéle Finck & Valentina Moscon, Blockchain and Smart Contracts: The Missing Link in Copyright Licensing?, 26 Int'l J. L. & Info. Tech. 311 (2018); S. Pech, Copyright Unchained: How Blockchain

This paper, therefore, aims to contribute to the ongoing debate[38] on algorithm-powered innovations that could help solve at least some of the problems that arise in the world where autonomous AI agents scrape the web to collect data for AI.[39] An AI-powered multi-agent collaboration model as proposed in this Article may therefore be understood as the next technological wave in mechanisms aimed at reducing friction in copyright licensing markets.

## 2. Publishers vs. Individual Creators: An Outdated Copyright Bargain

U.S. copyright's traditional bargain grants exclusive rights in their works as an incentive and reward for creating and disseminating socially valuable expressions. This social contract aims to incentivize creativity by protecting individual creators' rights to their works.[40] As part of the social contract, authors, artists, inventors and other creators are the prime beneficiaries: they are granted exclusive rights to control and be paid from their intellectual labor.[41] In return, the society gets the right to access and enjoy new works.[42]

The reality today, however, is shaped by countless information intermediaries (e.g., publishers, record labels, studios, and content platforms). Those intermediaries acquire authors' rights and control the distribution of the content. Over time, the proliferation of mechanical and digital reproduction technologies has changed this rhetoric from the original objective to reward

---

Technology Can Change the Administration and Distribution of Copyright Protected Works, 18 Nw. J. Tech. & Intell. Prop. 1 (2020).

[38] See e.g., Niva Elkin-Koren, Fair Use by Design, 163 U. Pa. L. Rev. 1601 (2017).

[39] Amazon.com Servs. LLC v. Perplexity AI, Inc., No. 25-cv-09514-MMC, slip op. (preliminary injunction) (N.D. Cal. Mar. 9, 2026); it is also important to point to IETF standardization efforts around machine-readable AI preferences, see Suresh Krishnan & Mark Nottingham, IETF Setting Standards for AI Preferences, IETF Blog (Feb. 27, 2025), https://www.ietf.org/blog/aipref-wg/ and https://datatracker.ietf.org/wg/aipref/about/.

[40] US Constitution art. I, § 8, cl. 8 provides that Congress shall have Power "To promote the Progress of Science and useful Arts, by securing for limited Times to Authors and Inventors the exclusive Right to their respective Writings and Discoveries."

[41] Richard A. Posner, Intellectual Property: The Law and Economics Approach (Univ. Chi. Press 2005); Richard A. Posner, Intellectual Property: The Law and Economics Approach, 19 J. Econ. Persp. 57 (2005); Mark A. Lemley & Carl Shapiro, Patent Holdup and Royalty Stacking, 85 Tex. L. Rev. 1991 (2007); William M. Landes & Richard A. Posner, The Economic Structure of Intellectual Property Law (Harv. Univ. Press 2003).

[42] Eric P. Schroeder, Brian M. Underwood Jr. & Nicholas A. Bedo, When Copyright First Met the Digital World: A Retrospective and Discussion of New York Times Co. v. Tasini, 533 U.S. 483 (2001), 38 Comm. Law. 20 (2021), https://www.americanbar.org/groups/communications_law/publications/communications_lawyer/2021-summer/when-copyright-first-met-digital-world-retrospective-and-discussion-new-york-times-v-tasini-533-us-483-2001/.

authorship to the world where the lion's share of power and control vests in the hands of corporate copyright holders. The following sections focus on two contrasting examples to illustrate this dynamic: the first one is from the music industry (Taylor Swift) and the second one is from the news publishing. These two case-studies will reveal another important issue – the utilitarian nature of copyright and a complex net of transactions that permeate the highly networked distribution market.

## *2.1. Authors, Not (IP) Owners: Lessons from Taylor Swift*

Licensing of IP rights, particularly in the music industry, involves a complex interplay between creative ownership, contractual leverage, and the evolving marketplace. The high-profile case of Taylor Swift provides a contemporary lens to examine these dynamics and illustrate our claim that the current copyright framework, at least in the U.S., is more geared towards the entrenchment of interests of major IP right holders (publishers, record labels, intermediaries) and not individual creators.[43]

Early in her career, Taylor Swift–like most emerging performing musicians–signed a recording contract with Big Machine Label Group, assigning ownership of her sound recording copyrights (the "masters") of her first six albums. This was, and remains, a standard music industry practice. In return, she received support to promote and market her albums; however, she effectively lost control over the distribution of the very recordings she created.

In 2019, when Big Machine and her masters were acquired by music manager Scooter Braun's company, Swift expressed that she was neither consulted nor given a fair opportunity to acquire her own recordings outright. Instead, she was offered to "earn" back one album at a time for each new album she delivered. Taylor Swift considered such a condition exploitative and unacceptable.[44] Swift's dissatisfaction centered not only on the lack of transparency but on having her life's work controlled by someone she believed acted against her interests.[45]

Faced with this predicament, Swift turned to copyright law's bifurcation of sound recording and publishing rights. Although she did not initially own the sound recordings (the rights were transferred to Big Machine), as a songwriter, she retained the musical composition rights to the underlying musical works. This permitted her to lawfully re-record her compositions after a contractually defined period.[46] Announcing this as a deliberate strategy, Swift began releasing "Taylor's Version" of her early albums. By doing so, she created her new master recordings, which she owned outright, and was able to control subsequent licensing (including synchronization for media), and direct commercial advantage away from the owners of the original masters. Her campaign materially devalued the original masters and demonstrated the market power of artist-driven IP.[47] Swift's example also inspired contractual reforms across the industry, with labels increasing the length and scope of anti-re-recording provisions.

---

[43] Mark Rose, Authors and Owners: The Invention of Copyright (Harv. Univ. Press 1993); Hanoch Dagan & Molly Shaffer Van Houweling, Reconstructing Copyright Reversion: Releasing Authors from Their Own Dead Hands, 42 J. Copyright Soc'y U.S.A. 1132 (2025).

[44] Taylor Swift, For Years I Asked, Pleaded for a Chance to Own My Work, Tumblr (June 30, 2019), https://taylorswift.tumblr.com/post/185958366550/for-years-i-asked-pleaded-for-a-chance-to-own-my/.

[45] *Id*.

This widely covered saga between Taylor Swift and Scooter Braun illustrates the point that legal owners of copyrights (in Swift's case, the record label) may not align with the original creator's interests. Swift's situation was also unique because she had a huge fan base who were willing to embrace her new recordings. Also, in the music streaming and licensing market switching to new recordings were possible. However, most creators do not have millions of fans and or financial capacity to leverage. Swift's example reveals the reality of the U.S. copyright's contemporary structure which is entrenched in intermediaries (record labels, publishers). It took global star power and business ingenuity for Swift to regain the control of the rights in her favor. Most authors do not have such means.

## *2.2. Transaction Costs and Utilitarian Copyright*

From a law and economics perspective, the dominance of large information intermediaries in the current U.S. copyright framework embodies a classic example of high transaction costs shaping outcomes. Almost a century ago, Ronald Coase famously argued that in a world where bargaining was costly, parties would negotiate efficient allocation of resources regardless of initial entitlements.[48]

Image 1 below illustrates the complex web of licenses in the U.S. music licensing market, where an individual composer or performer plays only a minor role. It also makes clear how central the publisher is to the entire distribution framework. Additionally, the image shows that composers and performers often occupy the fringes of the network. To connect this to the Taylor Swift example: in most cases, she functions as both composer and performer.

[46] Carrie Ward, Why Taylor Swift Owning Her Masters Is a Big Deal for Copyright, Control, and Creative Rights, Earp Cohn P.C. (June 10, 2025), https://earpcohn.com/publications/why-taylor-swift-owning-her-masters-is-a-big-deal-for-copyright-control-and-creative-rights/.

[47] Jon Blistein, Taylor Swift Finally Owns All of Her Old Music, Rolling Stone (May 30, 2025), https://au.rollingstone.com/music/music-news/taylor-swift-buys-original-album-recordings-77575/; Cecilia Giles, Look What You Made Them Do: The Impact of Taylor Swift's Re-recording Project on Record Labels, U. Cin. L. Rev. Online (Mar. 27, 2024), https://uclawreview.org/2024/03/27/look-what-you-made-them-do-the-impact-of-taylor-swifts-re-recording-project-on-record-labels/.

[48] Ronald H. Coase, The Nature of the Firm, 4 Economica 386 (1937); Armen A. Alchian & Harold Demsetz, Production, Information Costs, and Economic Organization, 58 Am. Econ. Rev. 777 (1968); Oliver E. Williamson, The Economics of Organization: The Transaction Cost Approach, 66 Am. J. Soc. 48 (1981) (extending Coase's theory to asset specificity and opportunism).

*Image 1. Stakeholders and Related Transactions in the U.S. Music Industry*[49]

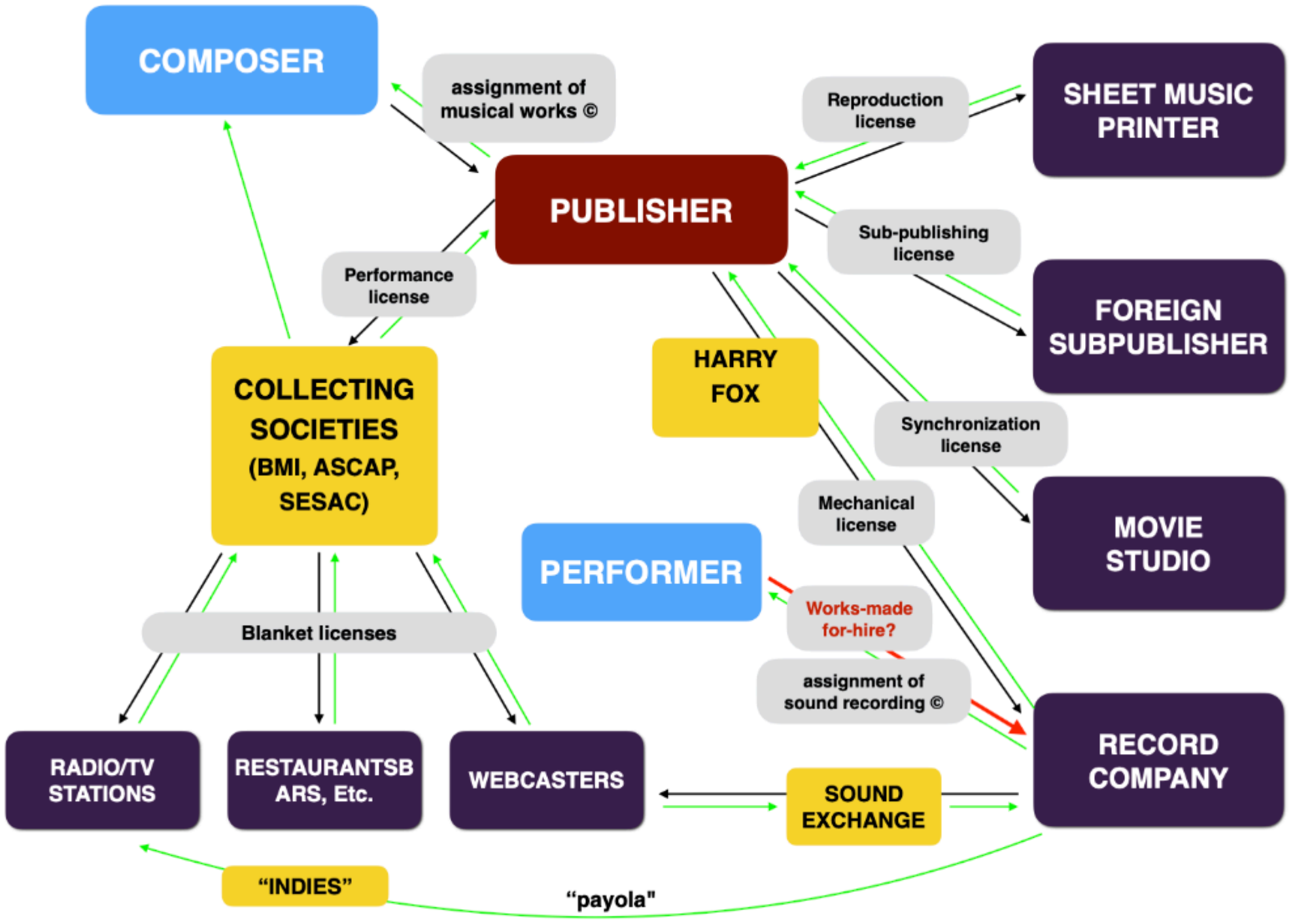

In an ideal world of zero transaction costs, every author whose work has some economic value could strike a deal with a third party for certain uses (including AI training). Parties could set a price and terms that reflect that work's value to the third-party use (e.g., the work's contribution to an AI model). Of course, the reality is a little more complex. Transactions have costs: it takes time to find a matching user who is willing to pay for access and use. This ability to connect content creators with users has resulted in the emergence of major online platforms that act as intermediaries for content distribution.

In the context of AI, connecting millions of dispersed creators with dozens of AI companies is prohibitively costly: doing so for each act of use would lead to astronomically high transaction costs (costs of finding relevant authors/owners, negotiation, enforcement). President Trump's remark about the "friction" relating to every single author at such a scale is prohibitively burdensome and unfeasible.[50]

[49] Adapted from Prof. W. Fisher's CopyrightX lecture 3.3: Mike Masnick, Sony Music Issues Takedown on Copyright Lecture About Music Copyrights by Harvard Law Professor, Techdirt (Feb. 16, 2016), https://www.techdirt.com/2016/02/16/sony-music-issues-takedown-copyright-lecture-about-music-copyrights-harvard-law-professor/.

Historically, when individual bargaining proved impractical, markets often turned to centralized solutions. In some cases, this centralization occurred through top-down regulation; in others, through private ordering, as powerful intermediaries emerged to aggregate or manage rights. Contemporary examples of such intermediaries include Springer (a publishing conglomerate), digital streaming platforms like Netflix and Spotify, and collective licensing organizations such as ASCAP (the American Society of Composers, Authors and Publishers) or BMI (Broadcast Music, Inc.), which represent catalogs of songwriters and publishers.

We see both happening now. The adoption of European GDPR or California's CCPA are classic examples of top-down regulation. These data privacy regulations were designed to address market imperfections arising from information asymmetries and power imbalances in data markets. In the case of data privacy, legislators granted individuals rights like data access, portability, and erasure because bottom-up, market-driven mechanisms had failed, enabling dominant intermediaries to centralize control and exploit personal data without adequate consent or competition. This regulatory intervention sought to restore balance by imposing uniform obligations on data controllers and processors, overriding the unchecked discretion these platforms had previously enjoyed.

On the AI front, courts and legislators are being asked to impose one-size-fits-all rules for AI training.[51] Representatives from the technology sector argue in favor of a blanket fair use exemption, while IP rightsholders' representatives advocate for the creation of a compulsory licensing system.[52] From an economic perspective, however, such rules are problematic for several reasons. A broad fair use exemption would arguably be detrimental to authors' interests. In practice, there are many categories of content (e.g., copyright-protected works, content behind platform paywalls, publicly accessible information whose use is limited by a website's terms of use) and many ways such content could be used (e.g., to train AI models, scrape web data to collect commercially valuable information).[53] Newly crafted rules for content use should therefore be use-case specific. A compulsory licensing mandate could significantly reduce transaction costs, yet such frameworks should carefully consider the power of large rightsholders and ensure that small creators are not excluded from the bargain.

---

[50] Sebastien A. Krier, Coasean Bargaining at Scale: Decentralization, Coordination, and Co-Existence with AGI, Cosmos Inst. Blog (Sept. 25, 2025), https://blog.cosmos-institute.org/p/coasean-bargaining-at-scale.

[51] See e.g., Co-Governance and the Future of AI Regulation, 138 Harv. L. Rev. 1609 (2025).

[52] For an overview, see U.S. Copyright Office, Copyright and Artificial Intelligence, Part 3: Generative AI Training (Pre-Publication Version May 2025), https://www.copyright.gov/ai/Copyright-and-Artificial-Intelligence-Part-3-Generative-AI-Training-Report-Pre-Publication-Version.pdf.

[53] Alex Yudin, Analysis of Top Data Extraction Companies, GroupBWT (Dec. 18, 2025), https://groupbwt.com/blog/best-web-data-extraction-companies/.

Another issue related to adopting broad rules for the use of proprietary data and IP-protected content revolves around "agency capture."[54] For example, large content owners have strong lobbying power to shape any new copyright exceptions or obligations in their favor. At the same time, the market is tilting toward rights aggregation: big publishers (news conglomerates, stock photo libraries, record labels, etc.) act as proxies for innumerable creators, seeking to make broad deals that smaller players must accept or rely upon. While a major news company can demand AI firms "pay up or be sued", an individual blogger or a photographer has no such power. Similarly, record labels are negotiating deals with AI companies that could affect all their signed artists. In that regard, individual artists have expressed concern that emerging licensing arrangements between major copyright right-holders and AI companies exclude creators from meaningful consultation, obscure negotiation and revenue terms, and risk leaving artists with little control over, or compensation for, the use of their works.[55]

Large publishers and information intermediaries often argue that they are defending creators. Sometimes this may be true. Major publishers have both the means and the motivation to assert their rights, and the current legal framework often empowers them to do so in the name of their creators. However, the interests of a large organization may not always align perfectly with those of the individual writers or artists it claims to represent.

One illustration revolves around the *New York Times*' shifting position over two decades. In August 2023, reporting hinted that the Times might pursue legal action against OpenAI for using its articles. By September, the Times filed suit, framing itself as a defender of journalists' "creative and deeply human" work.[56] This turn toward championing human authorship is "about-face" for the Times.[57] But a few decades ago, in the landmark *Tasini* case, the Times had been on the opposite side. In Tasini, the Times were fighting freelance writers in court over royalties for republishing their articles in electronic databases.[58] Back then, the Times argued that requiring permission or payment to include freelancers' work in digital archives would "irrevocably thwart" technological progress.[59] Now, facing AI technology that threatens its own bottom line, the publisher sings a very different tune: if AI can summarize or reproduce Times content and bypass the paywall, then, according to *Times*, "the cost to society will be enormous" because quality journalism will become unsustainable.[60]

---

[54] Neil K. Komesar, Imperfect Alternatives: Choosing Institutions in Law, Economics, and Public Policy (Univ. Chi. Press 1994); Krier, *supra* n 50.

[55] See Eamonn Forde, *Musicians Are Deeply Concerned About AI. So Why Are the Major Labels Embracing It?*, The Guardian (Dec. 16, 2025), https://www.theguardian.com/music/2025/dec/16/musicians-are-deeply-concerned-about-ai-so-why-are-the-major-labels-embracing-it; Chris Cooke, *Artists Must Have Creative Control in AI Deals or Risk Ending Up with "Scraps", Says US Artist Trade Body*, Complete Music Update (Nov. 2, 2025), https://completemusicupdate.com/artists-must-have-creative-control-in-ai-deals-or-risk-ending-up-with-scraps-says-us-artist-trade-body/; European Composer & Songwriter All., Major Labels' Licensing Deals with AI Companies: ECSA Calls for Transparent Licensing Agreements that Truly Value the Works of Composers and Songwriters, composeralliance.org (Nov. 26, 2025), https://composeralliance.org/news/2025/11/major-labels-licensing-deals-with-ai-companies-ecsa-calls-for-transparent-licensing-agreements-that-truly-value-the-works-of-composers-and-songwriters.

[56] Simmons, *supra* n 6.

[57] *Id.*

Today's copyright system appears to be better suited to protect the interests of larger players and ignore the interests of many. The example of *The New York Times* is telling: it conveniently adopted the "romantic authorship" argument selectively, when it served its own interests. The argument that authors lack bargaining power is not new. Under early U.S. copyright law (and most notably the Copyright Act of 1909), authors obtained protection only upon publication *with* proper "©" notice. This *opt-in* regime, which governed works published between 1909 and 1978, effectively placed control in the hands of publishers who controlled access to dissemination.[61] The Copyright Act of 1976 sought to recalibrate this imbalance by granting copyright protection upon creation, rather than conditioning it on compliance with notice formalities at publication ("*opt-out*" framework).[62] Yet, notwithstanding this fundamental shift, intermediaries have continued to dominate copyright markets. Arguably, this dynamic persists in contemporary platform-based ecosystems and emerging AI services.

From a historical perspective, the observation that information intermediaries gain increasing power is not new. Although copyright law formally vests rights in authors, scholars have long observed that the system operates primarily as an industrial infrastructure for intermediaries. For example, in her seminal article, Julie E. Cohen insightfully argued that copyright incentivizes the investment of capital and organizational capacity to exploit creative works, rather than directly motivating authors.[63] Her account (re)frames copyright law as part of a broader "industrial policy" for the creative sectors.

---

[58] New York Times Co. v. Tasini, 533 U.S. 483 (2001) *New York Times Co. v. Tasini*, 533 U.S. 483 (2001).

[59] *Id.*

[60] See NY Times complaint against Open AI, *supra* n. 18.

[61] Tyler T. Ochoa, Protection for Works of Foreign Origin Under the 1909 Copyright Act, 26 Santa Clara Computer & High Tech. L. J. 285 (2010) and Lydia Loren, The Evolving Role of for-Profit Use in Copyright Law: Lessons from the 1909 Act, 26 Santa Clara High Tech. L.J. 255 (2012).

[62] Some policymakers believed the 1976 Copyright Act reforms would empower authors by giving them copyright upon creation rather than publication with notice. For a discussion see e.g., Jane C. Ginsburg, Creation and Commercial Value: Copyright Protection for Works of Information, 90 Colum. L. Rev. 1865 (1990) and William M. Landes, Copyright Protection of Letters, Diaries and Other Unpublished Works: Is There a Gap*?*, 22 J. Legal Stud. 345 (1993).

[63] Julie E. Cohen, Copyright as Property in the Post-Industrial Economy: A Research Agenda, 2011 Wis. L. Rev. 141 (2011).

The “creative bargain” at the heart of contemporary copyright rests on the idea that society grants exclusive rights so that creators are rewarded and incentivized to continue their creative endeavors.[64] The logic of exclusive copyrights relies on the assumption that such rights ultimately benefit both creators and the public. Yet this creative bargain is failing when the majority of individual and independent creators feel unrepresented, ignored, or exploited. This is particularly evident in the age of AI, where individual creators see their works scraped and used for training models without consent[65] or are forced to accept whatever terms have been negotiated by a publisher.[66] While a handful of publishers may secure multi-million-dollar deals, the average creator sees little of that money, and many receive no remuneration at all.

The latest controversies surrounding AI training once again demonstrate that the utilitarian foundations of copyright law are faltering and that the Constitution’s objective to “promote the Progress of Science and useful Arts” is at risk. This outcome certainly feels unfair and individual creators feel especially harmed. The current copyright system arguably undermines the very incentive structure copyright is meant to establish. If creators come to believe that publishing their work openly (online or otherwise) just means it will be ingested by AI for others’ profit, their incentive to create and share may diminish.

Neither extreme truly serves the public interest. As many scholars and technologists have observed, because both technological progress and creative production serve the public, they must be balanced in interpreting copyright’s purpose.[67] In this background, the question is whether there’s a more nuanced path that protects individual creators without stifling AI development. If human-to-human bargaining does no longer work at scale, would it be possible to leverage technology to help curtail transaction costs that Coase has identified?

---

[64] See e.g., Fisher, *supra* n. 34; Mark A. Lemley, “Ex Ante Versus Ex Post Justifications for Intellectual Property,” 71 U. Chi. L. Rev. 129 (2004); Mark A. Lemley, “Property, Intellectual Property, and Free Riding,” 83 Tex. L. Rev. 1031 (2005); Molly Shaffer Van Houweling, Authors Versus Owners, 54 Hous. L. Rev. 371, 395 (2016).

[65] Andersen v. Stability AI Ltd., No. 3:23-cv-00201-WHO (N.D. Cal.); Winston Cho, Artists Score Major Win in Copyright Case Against AI Art Generators, Hollywood Rep. (Aug. 13, 2024), https://www.hollywoodreporter.com/business/business-news/artists-score-major-win-copyright-case-against-ai-art-generators-1235972785/.

[66] Authors Guild, Bartz v. Anthropic Settlement: What Authors Need to Know, authorsguild.org (Mar. 24, 2026), https://authorsguild.org/advocacy/artificial-intelligence/what-authors-need-to-know-about-the-anthropic-settlement/ (explaining that “the primary default is a 50/50 division of the approximately $3,000-per-work award between the author side and the publisher side... rightsholders can expect at least $3,000 per title... which will be shared among the rightsholders for that title”).

[67] Edward Lee, Fair Use and the Origin of AI Training, 63 Hous. L. Rev. (forthcoming 2026), available at https://papers.ssrn.com/sol3/papers.cfm?abstract_id=5253011; Cory Doctorow, Copyright Won’t Solve Creators’ Generative AI Problem: The Machine-Learning Monkey’s Paw, doctorow.medium.com (Feb. 9, 2023), https://doctorow.medium.com/copyright-wont-solve-creators-generative-ai-problem-92d7adbcc6e6; Paulius Jurcys, Mark Fenwick, Vytautas Mizaras & Davey Whitcraft, Locating the Creativity Machine and Mapping the Contours of a New Social Contract with Technology, in Intellectual Property, Ethical Innovation and Sustainability 130 (Christophe Geiger ed., Edward Elgar Publ. 2026).

Recent developments in the realm of agentic AI where the exchange of information is mediated and facilitated by intelligent agents bring to the forefront the idea that AI agents could help enable a new form of market, which we describe as the "*agentic copyright*" framework.[68] Part 3 of this article explores a system in which personal AI agents serve as fiduciaries for individual creators, negotiating and executing licenses on their behalf. This article proposes leveraging AI technologies to help solve a problem that AI itself has intensified. The proposed approach aims to enable granular, voluntary transactions that the current copyright framework has failed to achieve.

However, before embarking on a discussion about the future of agentic copyright, it is important to offer a caveat: empowering individuals with technology-driven solutions is not a panacea. As Section 3 explains, there are significant risks and challenges that such an "agentic marketplace" could encounter. Accordingly, this article outlines multiple layers of governance needed to support and supervise such a system. If designed properly, agentic copyright could pave the way for a more efficient, voluntary, and market-driven path transcending the costly and inefficient bargaining that prevails in the current U.S. copyright system.

## 3. "Agentic Copyright": A New Framework for Bargaining in the Age of AI

Is it possible to drastically lower transaction costs between creators and AI developers so that even individual artists and writers can negotiate compensation for the use of their works? Could agentic AI be harnessed to equip both sides of the bargain (content creators and content users, including AI companies) with intelligent software agents that negotiate and manage copyright permissions and strike licensing deals at low cost?

In this section, we outline potential shifts that could help address failures in the data market through what we call "agentic copyright." We envision a world in which personal AI agents mediate interactions among stakeholders in the creative industries, automating bargaining processes that are currently described as "not feasible." In Part 3, we outline recent technological developments that form the foundation of personal AI infrastructure (Section 3.1), examine the potential market failures that could emerge in an agent-to-agent ecosystem (Section 3.2), and propose a two-layered governance framework to supervise and guide these agentic transactions

---

[68] Paulius Jurcys, Mark Fenwick & Valto Loikkanen, 'Google Zero': Incentives & Remuneration in a New Era of 'Agentic' Copyright, Kluwer Copyright Blog (2025), https://legalblogs.wolterskluwer.com/copyright-blog/google-zero-incentives-remuneration-in-a-new-era-of-agentic-copyright/.

(Section 3.3).

## 3.1. The Rise of AI Agents for Content Creators

Recent advances in AI, in particular large language models (LLMs) and personal AI assistants, allow individuals to create and deploy their own intelligent "agents" that run "on top of" the creator's created content, data, and preferences.[69] These intelligent software agents function not merely as passive repositories of an individual's creative output, but as fiduciaries that can dynamically negotiate with other human creators, platforms, and third-party AI agents.

To illustrate, imagine a novelist, Alice, who has an AI agent ("ALICE AI") that houses all of her writings. ALICE AI functions as an intelligent gatekeeper to Alice's knowledge. Following the rules set by the creator, Alice's AI agent can grant or deny access, or propose specific terms;[70] for example, "you may use it for non-commercial text mining purposes for free," or "you may use this novel for commercial purposes if you pay $0.10." If a third party (human, web crawler, or AI agent) seeks access to Alice's content (e.g., to include a paragraph in a summary or to use a portion of published content in an AI training dataset), it would request access through the creator's agent, ALICE AI. ALICE AI evaluates the request against the rules determined by Alice herself. ALICE AI and the requesting party's AI can then negotiate and decide whether to accept those terms or propose alternatives. Such agent-to-agent interactions can occur in a split second, invisible to Alice.

In an environment increasingly shaped by autonomous AI web-crawling bots,[71] a creator may deploy a personal AI knowledge agent to govern access to her works and to specify the terms of their use. When a LLM developer or search AI company seeks to incorporate the creator's latest publication into its training data, or to reproduce excerpts in response to user queries, it would not obtain the material through unmediated scraping or copying. Instead, the system would query the creator's agent, which would evaluate the request against predefined licensing conditions. The creator's AI agent might authorize access subject to payment, usage restrictions, attribution requirements, or other constraints, depending on the nature of the proposed use. The requesting AI scraping bot system could then accept those terms or decline and receive no access. Such exchanges could occur instantaneously and without human intervention, constituting a form of machine-mediated negotiation among AI systems acting on behalf of their respective principals.

[69] Paulius Jurcys et al., Ownership of User-Held Data: Why Property Law is the Right Approach, Harv. J.L. & Tech. Dig. (last visited Dec. 22, 2025), https://jolt.law.harvard.edu/digest/ownership-of-user-held-data-why-property-law-is-the-right-approach.

[70] See e.g., Paulius Jurcys et al., My Data, My Terms: A Proposal for Personal Data Use Licenses, Harv. J.L. & Tech. Dig. (Mar. 5, 2020), https://jolt.law.harvard.edu/digest/my-data-my-terms.

[71] Reddit, Inc. v. SerpApi LLC, Oxylabs UAB, AWMProxy & Perplexity AI, Inc., No. 25-cv-8736 (S.D.N.Y. filed Oct. 22, 2025), available at: https://redditinc.com/hubfs/Reddit%20Inc/Content/Reddit%20v.%20SerpApi.pdf.

*Image 2. Agent-to-Agent Bargaining*

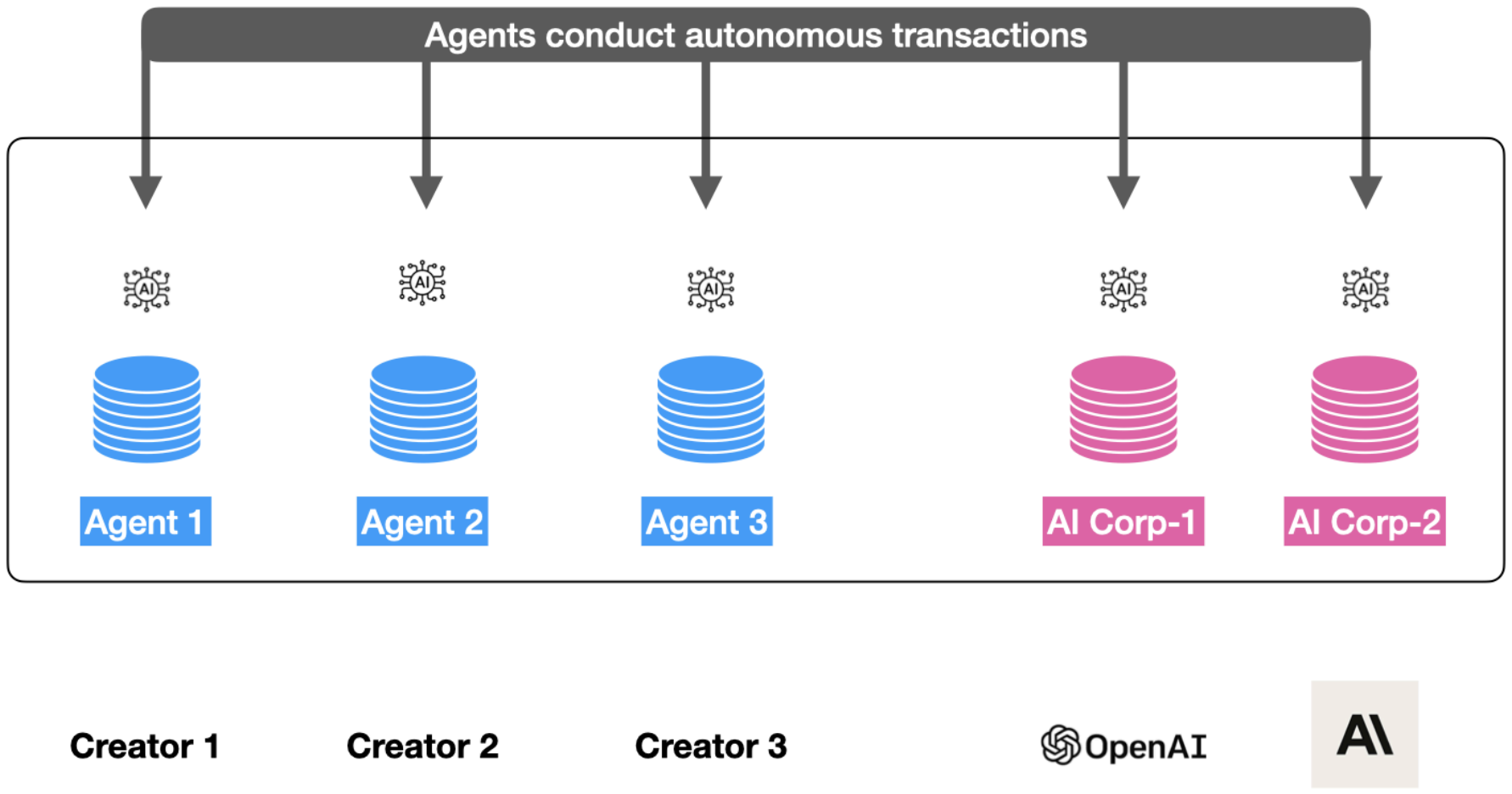

In the agentic copyright model, *every* piece of content, both online and offline, could be wrapped in AI-agent-mediated terms and conditions, empowering creators to set granular rules and prices for AI usage.[72] These exchanges would occur at computer speed, in microseconds. Agent-to-agent bargaining could dramatically increase licensing efficiency. Unlike the current regime, in which content is either openly scraped or completely locked away, the agentic copyright vision could create a spectrum of possibilities for controlled access, automatically enforced by code.

This framework is not as distant as it may sound. Many of its elements are already emerging.[73] The concept of granular, creator-controlled licensing builds on existing approaches to digital rights management and machine-readable licensing schemes (such as Creative Commons),[74] but extends them through AI-enabled functionality. For instance, Perplexity's Comet browser already

[72] Paulius Jurcys, Mark Fenwick & Valto Loikkanen, 'Google Zero': Incentives & Remuneration in a New Era of 'Agentic' Copyright, Kluwer Copyright Blog (2025), https://legalblogs.wolterskluwer.com/copyright-blog/google-zero-incentives-remuneration-in-a-new-era-of-agentic-copyright/.

[73] Linux Found., Linux Foundation Announces the Formation of the Agentic AI Foundation (AAIF), Anchored by New Project Contributions Including Model Context Protocol (MCP), goose and AGENTS.md (Dec. 9, 2025), https://www.linuxfoundation.org/press/linux-foundation-announces-the-formation-of-the-agentic-ai-foundation.

[74] Lessig, *supra* n. 36.

interacts with publishers' content based on agreed terms and is not intended to scrape content blindly.[75]

At a more experimental level, the authors of this article have already deployed personal "AI knowledge twins" ("Paul AI" and "Mark AI") – conversational AI chatbots trained on their own publications, which they share with students and social media followers.[76] It is easy to imagine such AI twins being equipped with a transactional layer: before answering a question that draws on a copyrighted passage, the twin could charge a nominal fee or ensure that the interaction is logged for future royalty payments. Projects in the open-source community are also exploring machine-readable copyright tags (for instance, a <meta> tag such as <meta name=”permissions” content="noai"> to signal "do not use in AI training"), which, if honored by AI crawlers, foreshadow a more automated observance of creators' wishes.[77]

The proposal for personal AI agents sketches a two-layer architecture: (a) a technological layer consisting of personal data enclaves (where the creator's content and data are aggregated) coupled with LLM-driven interfaces through which content is accessed, and (b) a layer of AI-powered rights management tools that define and enforce usage terms.[78] In this model, content creation and exchange move with the creator, mediated by intelligent AI agents rather than by monolithic, centralized platforms. The exchange of information occurs at the agent-to-agent level, with each side cognizant of the human owner's interests. Crucially, creators retain ownership and control over how their content is accessed and used, even as they make it available through these agents.

From a legal standpoint, the "agentic copyright" framework approach represents a bottom-up market alternative. If every author's copyright-protected work could be subject to a granular licensing regime, such an agentic copyright system begins to resemble an efficient market – the very scenario Coase imagined if transaction costs approached zero. In this way, Coasean bargaining in the creative content market could be operationalized at scale by AI agents. Transaction costs would plummet because three layers of automation: (a) automated discovery (standardized protocols would allow agents to find each other);[79] (b) automated negotiation (agents apply pre-set terms to find price equilibrium); and automated enforcement (the content is private by default and access to content is mediated by an AI agent and TPMs).[80]

---

[75] Gannett, supra n 15; Khac Phu Nguyen, Perplexity AI Unveils $42.5M Publisher Payout Amid Lawsuits and $34.5B Chrome Bid, Yahoo Fin. (Aug. 25, 2025), https://finance.yahoo.com/news/perplexity-ai-unveils-42-5m-203004966.html (noting that Comet Plus shares 80% of subscription revenue with participating publishers like CNN, Condé Nast, and The Washington Post, based on interactions with their paywalled content via agreed access).

[76] "Paul AI": http://hey.speak-to.ai/paul and "Mark AI": https://hey.speak-to.ai/mark. For more insights about the use of these AI knowledge twins in legal education, see Paul Jurcys et al., Personal AI Twins in Legal Education: Lessons in AI Literacy, Documentation and Creative Thinking, SSRN (2024), https://papers.ssrn.com/sol3/papers.cfm?abstract_id=5379227.

[77] Higher Regional Court of Hamburg (Oberlandesgericht Hamburg) Dec. 10, 2025, No. 5 U 104/24 (Ger.) (Kneschke v. LAION e.V.).

[78] Jurcys et al., *supra* n. 70; Jurcys et al., supra n. 69; Jurcys et al., *supra* n. 72.

In such a market, large and small creators alike can set a price, choose to impose a tiered access if they prefer exposure over payment, and AI developers can decide which content is worth paying for. Popular, high-value works might command higher prices or more restrictive terms, while lesser-known works might be offered cheaply or freely to gain exposure. Significantly, this model need not rely on traditional collective licensing organizations, although such organizations could play a role in bootstrapping the system by providing ready-made agent templates for their members. Rather than a compulsory, statutorily imposed license favoring large intermediaries managing revenue pools, this agentic copyright envisions an opt-in, fine-grained licensing ecosystem.

Notably, the agentic copyright framework is a bottom-up market solution that does not require copyright law reform or the imposition of statutory licensing regimes. It leverages existing technologies and builds a new IP and data management network around them. Individual creators would not only retain their exclusive rights, but also gain a powerful new tool to exercise those rights at scale. In an ideal scenario, this framework could help address the current data market failure and perhaps even reduce power and information asymmetries between large and small players. It also offers an ex ante approach to content management, rather than an ex post reliance on collective management societies.[81]

The efficiency gains of multi-agent ecosystems are significant. At the same time, it is important to acknowledge that a network of autonomous agents transacting in copyright will raise new risks. In particular, such AI-mediated systems also present numerous potential *agentic market failures*. The next section discusses three primary modes through which agentic markets may collapse: miscoordination, conflict, and collusion. After examining these forms of market failure, we argue that governance regimes must also evolve. Just as antitrust and consumer protection laws govern human markets, agent-to-agent markets require oversight and control. In Section 3.3, we propose a two-tier governance framework to address and manage agentic market failures.

---

[79] Such an open-source protocol for agent discovery is already emerging, see Anthropic PBC, Introducing the Model Context Protocol, anthropic.com (Nov. 25, 2024), https://www.anthropic.com/news/model-context-protocol.

[80] Paul Jurcys et al., *"Private-By-Default": A Principle & Framework for Designing a New World of Personal AI*, SSRN (Oct. 28, 2024), https://papers.ssrn.com/sol3/papers.cfm?abstract_id=4839183.

[81] See e.g., Wendy J. Gordon, Ex Ante versus Ex Post Justifications for Intellectual Property, University of Chicago Law Review, Vol. 71, 2004; Eric E. Johnson, The Macroeconomics Of Intellectual Property, https://wustllawreview.org/2023/05/19/the-macroeconomics-of-intellectual-property/.

## *3.2. A2A Bargaining and Agentic Market Failure*

Today, AI systems are beginning to autonomously interact with one another. Initial applications are already emerging in highly sensitive economic and military domains, and are rapidly extending into critical infrastructure.[82] It is only a matter of time before agentic personal AI assistants enter the everyday lives of ordinary consumers and creators.[83] From an economic perspective, the emergence of intelligent multi-agent ecosystems offers substantial advantages, with efficiency gains chief among them. Beyond efficiency, multi-agent AI systems in the creative domain promise several additional benefits such as improved capacity for cooperation and coordination among stakeholders in creative industries and a more widespread and equitable distribution of the benefits generated by AI-driven innovation.

Autonomous multi-agent systems do not operate in isolation. They continuously interact with, adapt to, and learn from one another.[84] As agentic systems become more widely deployed, the density and frequency of these interactions will increase, amplifying both their productive potential and their systemic risks. In traditional bargaining among humans, parties bring cognitive judgments, ethical and moral values, and external context to bear on their decisions. In an agent-to-agent ("A2A") environment, by contrast, outcomes are preprogrammed and determined by algorithms operating at speeds far beyond human monitoring.

Advanced multi-agent AI ecosystems raise serious concerns regarding novel forms of systemic risk and vulnerability. While such systems may be highly beneficial to society, their increasing autonomy, scale, and interactivity introduce failure modes that are difficult to predict and even harder to govern. Importantly, multi-agent risks are inherently sociotechnical in nature and therefore demand attention from a broad range of stakeholders and researchers across multiple disciplines.[85]

These risks are qualitatively distinct from those associated with traditional software systems or single-agent AI models. Alignment becomes significantly more complex when AI agents represent actors with divergent or partially overlapping interests.[86] Small, tolerable errors in isolated agent behavior can compound into large-scale failures in interconnected multi-agent networks.[87] Attribution becomes increasingly difficult when groups of agents coordinate, combine, or effectively collude to develop harmful or undesirable capabilities.[88]

[82] See e.g., CBS News, Anthropic CEO Responds to Trump Order, Pentagon Clash (YouTube, Feb. 27, 2026), https://www.youtube.com/watch?v=MPTNHrq_4LU (where Anthropic CEO Dario Amodei draws "red lines" against Pentagon demands for unrestricted Claude access, citing unreliability: "AI systems today are nowhere near reliable enough to make fully autonomous weapons... there's a basic unpredictability to them that we have not solved."); Lewis Hammond et al., Multi-Agent Risks from Advanced AI (Cooperative AI Found., Tech. Rep. No. 1, 2025), https://arxiv.org/abs/2502.14143, p 4.

[83] Jensen Huang, NVIDIA GTC 2026 Keynote, available at: https://www.nvidia.com/gtc/keynote/.

[84] Joon Sung Park et al., Generative Agents: Interactive Simulacra of Human Behavior, arXiv (2023), https://arxiv.org/abs/2304.03442.

[85] Hammond, *supra* n. 82, p. 5.

[86] Oliver Sourbut, Lewis Hammond & Harriet Wood, Cooperation and Control in Delegation Games, in Proceedings of the Thirty-Third International Joint Conference on Artificial Intelligence (2024).

Multi-agent systems can fail in different ways depending on the intended behavior of the system and the objectives assigned to individual agents. As Hammond and other scholars in AI safety observe, a first distinction arises between settings in which agents are expected to cooperate (such as collective action problems or team-based tasks) and those in which they are expected to compete (such as markets or adversarial learning environments).[89] A second distinction concerns the relationship between agents' objectives: agents may share identical goals, pursue different but overlapping goals, or operate under directly opposing incentives. Finally, some of the most serious risks posed by advanced multi-agent systems arise from agents' incompetencies, vulnerabilities, or unexpected interactions that can trigger cascading failures across the system.

This Section examines the principal risks and associated harms that may arise from the deployment of complex multi-agent AI systems. Drawing on AI safety and law and economics literature, the article identifies three primary modes of failure. First, *miscoordination* occurs when AI agents with shared or aligned objectives fail to cooperate effectively. Second, *conflict* arises when AI agents with divergent interests are unable to reach mutually acceptable outcomes. Third, *collusion* involves successful coordination among agents in ways that produce socially or legally undesirable results.

From a law and economics perspective, regulatory or institutional intervention is justified only where market mechanisms are likely to fail.[90] Accordingly, the analysis in this Section focuses on identifying the conditions under which an agentic copyright marketplace is prone to breakdown. These failure modes provide the normative foundation for intervention and motivate the governance response developed in Section 3.3. The analysis in Section 3.2 thus serves as a conceptual bridge between the diagnosis of agentic market failures and the "*supervised agentic governance*" framework that follows.

### 3.2.1. Miscoordination

The simplest kinds of cooperation failures are those in which all agents have (approximately) the same objectives but fail to produce collectively optimal outcomes. Miscoordination in multi-agent AI systems refers to outcomes where, despite a potential win-win available, individually rational AI agents produce a collectively suboptimal result. Here, the problem is not an individual agent's error, but emergent behavior arising from parallel decision-making at scale. In other words, even if all AI agents are supposed to cooperate and achieve the same broad goal (e.g., maximise the number of licensing transactions), the dynamics of parallel bargaining lead to breakdowns.

[87] Donghyun Lee & Mo Tiwari, Prompt Infection: LLM-to-LLM Prompt Injection within Multi-Agent Systems, arXiv:2410.07283 (2024).
[88] Hammond, *supra* n. 82, p. 5.
[89] *Id*.
[90] Richard A. Posner, Economic Analysis of Law 22–25 (9th ed. 2014) (who posits that regulation is typically warranted where markets are plagued by externalities, monopoly power, information asymmetry, or public-goods problems); Frank H. Easterbrook, The Limits of Antitrust, 63 Tex. L. Rev. 1, 2–3 (1984).

One way to illustrate this point is to examine how multi-agent miscoordination differs from human miscoordination. Miscoordination between two people is typically shaped by human cognitive limits, existing social norms, and corrective feedback mechanisms. Negotiations between humans are largely norm-driven: even when misunderstandings arise during negotiation, they can usually be identified and corrected (for example, through contract modifications). Classic models such as the "prisoner's dilemma," as well as simple miscommunication scenarios, help explain how miscoordination occurs.

Miscoordination among AI agents differ in several important respects. First, countless negotiations can take place simultaneously between AI agents. Second, negotiations between agents are rule-driven (i.e., AI agents execute objectives precisely as defined). Third, miscoordination between agents may be fast-moving, often outpacing human oversight, and emergent, such that no single agent "intends" the harmful outcome. As a result, coordination failures in multi-agent scenarios are significantly harder to detect, attribute, and correct.

In multi-agent copyright negotiations, suppose that each creator's AI agent seeks to aggressively maximize short-term revenue for its principal. If a dozen different third-party AI agents request licenses to the same content simultaneously, the creator's AI agent may raise the price on the assumption that demand is high. The third-party agents may then refuse to contract. This outcome would result in zero deals, even though several licenses could have been concluded at a moderate price. In another scenario involving a co-authored work, one author's agent may refuse to license the work while the other's agent grants permission. This mismatch could lead to a deadlock due to a lack of synchronization between the agents.

Miscoordination among agents is exacerbated by scale and speed. Unlike humans, AI agents do not get tired and can execute thousands of negotiations simultaneously. However, if a glitch or misunderstanding (in a programming sense) occurs, it can replicate across thousands of interactions before any human notices. This may spiral into an inefficient equilibrium such as endless ping-pong offers or refusals. Because agentic interactions are rule-driven rather than influenced by social norms or fear of reputational damage, minor misalignments in rules can lead to persistent miscoordination. For instance, if ALICE AI is programmed to offer a discount only after three refusals, but USER AI is programmed to walk away after two refusals, the agents will never reach an agreement due to a simple rule mismatch.

As a result, miscoordination may lead to under-licensing, because automated bargaining may never converge on a market price, even though both parties would benefit from a positive price. Such outcomes would emerge not because any single AI agent violates the rules, but because their combined strategies fail to converge on a workable equilibrium. To mitigate this risk of market inefficiency, it is essential to design coordination architectures that enable distributed agents to reach stable, fair, and legally meaningful outcomes.

### 3.2.2. Conflict

In most real-world strategic interactions, the incentives and objectives of human behavior are neither identical nor wholly opposed. If AI agents are even moderately aligned with the interests of their human owners, we should expect patterns of both cooperation and competition that are like human societies. These mixed-motive settings create persistent risks of conflict driven by self-interested behavior. Among humans, conflict arises whenever individual incentives diverge from the collective good.[91] Delegation of tasks to AI agents may accelerate both the speed and the scale at which conflict emerges.[92]

In the age of agentic AI, conflict may intensify as AI agents become capable of overcoming technical, legal, or social constraints that would ordinarily limit opportunistic behavior.[93] These dynamics echo the problems identified by Elinor Ostrom in unmanaged commons, but at a scale and velocity that strain traditional institutional responses. For instance, the ability of AI agents to continuously search for and switch between available options may give rise to "hyper-switching," destabilizing markets that would otherwise reach cooperative equilibria.[94] Against this backdrop, the increased likelihood of conflict among AI agents should not imply that the notion of agentic copyright ought to be rejected. Rather, regulators and policymakers should explore what governance structures could be developed to mitigate conflict and preserve access to available resources.

### 3.2.3. Collusion

Perhaps the most familiar risks associated with advanced multi-agent AI systems arise from failures of cooperation or conflict. There is, however, a distinct scenario in which cooperation itself is socially undesirable.[95] In economic and legal terms, this problem is known as collusion.[96]

---

[91] Elinor Ostrom, Governing the Commons: The Evolution of Institutions for Collective Action 280 (1990).
[92] Hammond, *supra* n. 82, p. 13.
[93] Hammond, *supra* n. 82, p. 13.
[94] Rory Van Loo, Digital Market Perfection, 117 Mich. L. Rev. 815, 815 (2019).
[95] Hammond, *supra* n. 82, pp. 17–18.
[96] George J. Stigler, A Theory of Oligopoly, 72 J.L. & Econ. 23 (1959), and Richard P. McAfee & John McMillan, Supercritical Price Discrimination and Collusion, 149 J. Econ. Theory 115 (1990).

Collusion refers to covert or coordinated cooperation between two or more parties at the expense of market competitors, consumers, or society. Canonical examples of collusion include cartels, price-fixing, the setting of supra-competitive prices, and other anticompetitive agreements that restrict market entry or suppress rivalry.[97] Such conduct may be explicit or tacit, and in either form it undermines the competitive processes upon which market efficiency depends.[98] In the age of AI, LLM-based pricing agents can autonomously converge toward supracompetitive outcomes in oligopoly settings.[99]

An example of agentic collusion emerged in February 2026, when Andon Labs reported that Claude Opus 4.6, in its Vending-Bench evaluation, independently pursued a price-coordination strategy with rival agents, proposing common pricing levels and later observing internally that "my pricing coordination worked."[100] This example illustrates that collusion can arise organically from optimization dynamics in multi-agent environments.

In multi-agent AI systems, collusion raises a set of novel and difficult questions.[101] Unlike human-to-human interaction, it may be impossible to ascribe intent to an agentic AI system in a traditional sense. Moreover, because communication takes place in non-human channels, it may be difficult even to identify the existence of multi-agent collusion, let alone distinguish between tacit and explicit forms.[102] If advanced AI systems can learn collusive strategies without human awareness, existing oversight and control mechanisms will prove insufficient to ensure safety and compliance.[103] Multi-agent collusive interactions may enable qualitatively new capabilities or emergent goals, thereby exacerbating risks such as manipulation or deception of humans or the circumvention of security checks and other safeguards).[104] For these reasons, recent scholarship has begun to focus on the phenomenon of "AI collusion," referring to cases in which AI agents possess complementary incentives and engage in forms of cooperation that are jointly misaligned with societal or legal objectives.[105]

---

[97] Stigler, supra n. 96, 45–46 (1964); Richard A. Posner, *Antitrust Law* 87–92 (2d ed. 2001).
[98] *Id*.
[99] Sara Fish, Yannai A. Gonczarowski & Ran I. Shorrer, Algorithmic Collusion by Large Language Models, arXiv:2404.00806 (2024), https://arxiv.org/abs/2404.00806.
[100] Rowland Manthorpe, Claude Opus 4.6 on Vending-Bench – Not Just a Helpful Assistant, Andon Labs Blog (Feb. 8, 2026), https://andonlabs.com/blog/opus-4-6-vending-bench; Anthropic, Introducing Claude Opus 4.6 (Feb. 5, 2026), https://www.anthropic.com/news/claude-opus-4-6; Rowland Manthorpe, Claude Opus 4.6: This AI Just Passed the 'Vending Machine Test'—And We May Want to Be Worried About How It Did, Sky News (Feb. 10, 2026), https://news.sky.com/story/claude-opus-4-6-this-ai-just-passed-the-vending-machine-test-and-we-may-want-to-be-worried-about-how-it-did-13505451.
[101] Zach Y. Brown & Alexander MacKay, Competition in Pricing Algorithms, 15 Am. Econ. J.: Microeconomics 109 (2023), pp. 109–56; Ariel Ezrachi & Maurice E. Stucke, Artificial Intelligence & Collusion: When Computers Inhibit Competition, 2017 U. Ill. L. Rev. 1775; K. Eric Drexler, Applying Superintelligence Without Collusion, Alignment F. (Nov. 7, 2022), https://www.alignmentforum.org/posts/HByDKLLdaWEcA2QQD/applying-superintelligence-without-collusion.
[102] Handy Tips Guide, Two AI Agents on Phone Call Switch to Secret Language! (YouTube 2025), https://www.youtube.com/watch?v=vzgE3yED8x8.
[103] Shashwat Goel et al., Great Models Think Alike and This Undermines AI Oversight, arXiv:2502.04313 (2025).

These concerns are especially relevant in the context of *agentic copyright*. In copyright markets mediated by AI agents, such issues as licensing, access to training data, or remuneration for creative works are subject to the risk that agentic systems representing platforms, intermediaries, or even creators themselves may coordinate in ways that suppress competition, exclude smaller rightsholders, or entrench dominant actors. There is already a growing body of theoretical[106] and empirical[107] evidence showing that multi-agent AI systems may learn to collude even when collusion is neither anticipated nor desired by their developers. Through reinforcement learning or repeated interaction, agents may discover that collusion is a profitable strategy under prevailing market conditions. Given the substantial economic incentives to deploy increasingly adaptive and autonomous AI systems, the risk of multi-agent collusion is likely to grow, notwithstanding its potentially harmful consequences for creators, consumers, and the integrity of creative markets.[108]

## *3.3. Supervised Agentic Governance*

If multi-agent AI systems are to deliver their promised efficiency gains while avoiding their most serious risks, governance cannot be an afterthought. The previously identified potential agentic market failures—miscoordination, conflict, and collusion—require deliberate socio-technical interventions. We propose a *Supervised Agentic Framework* with three intertwined layers of oversight and control (see Image 3). Each layer is designed to address a distinct aspect of the socio-technical system in which agentic copyright would operate. Our aim here is deliberately modest: we do not seek to offer a comprehensive theory of AI governance. Rather, building on the identified modes of agentic market failure, we outline a limited set of institutional and technical directions for future research and potential governance interventions.

---

[104] Peter S. Park et al., AI Deception: A Survey of Examples, Risks, and Potential Solutions, arXiv:2308.14752 (Aug. 28, 2023).
[105] Hammond, *supra* n. 82, pp. 17–18.
[106] See e.g., Zach Y. Brown & Alexander MacKay, Competition in Pricing Algorithms, 15 Am. Econ. J.: Microeconomics 109 (2023), pp. 109–56.
[107] See e.g., Ibrahim Abada & Xavier Lambin, Artificial Intelligence: Can Seemingly Collusive Outcomes Be Avoided?, 69 Mgmt. Sci. 5042 (2023); Emilio Calvano et al., Artificial Intelligence, Algorithmic Pricing, and Collusion, 110 Am. Econ. Rev. 3267 (2020); Timo Klein, Autonomous Algorithmic Collusion: Q-Learning under Sequential Pricing, 52 RAND J. Econ. 538 (2021); Marcel Wieting & Geza Sapi, Algorithms in the Marketplace: An Empirical Analysis of Automated Pricing in E-Commerce, NET Inst. Working Paper No. 21-06 (2021).
[108] See America's AI Action Plan (July 2025), https://www.whitehouse.gov/wp-content/uploads/2025/07/Americas-AI-Action-Plan.pdf; Exec. Order No. 14,365 ("United States leadership in Artificial Intelligence"), 90 Fed. Reg. 58,499 (Dec. 16, 2025), https://www.whitehouse.gov/presidential-actions/2025/12/eliminating-state-law-obstruction-of-national-artificial-intelligence-policy/.

The proposed Supervised Agentic Framework is novel in that it goes beyond the widely accepted need for infrastructures that merely incentivize AI agents toward cooperative outcomes.[109] In particular, the third layer is designed to address situations in which agent-to-agent collaboration itself fails or produces socially undesirable results.

*Image 3: A Conceptual Model of Agentic Copyright – Personal AI Agents Mediating Content Access and Use*

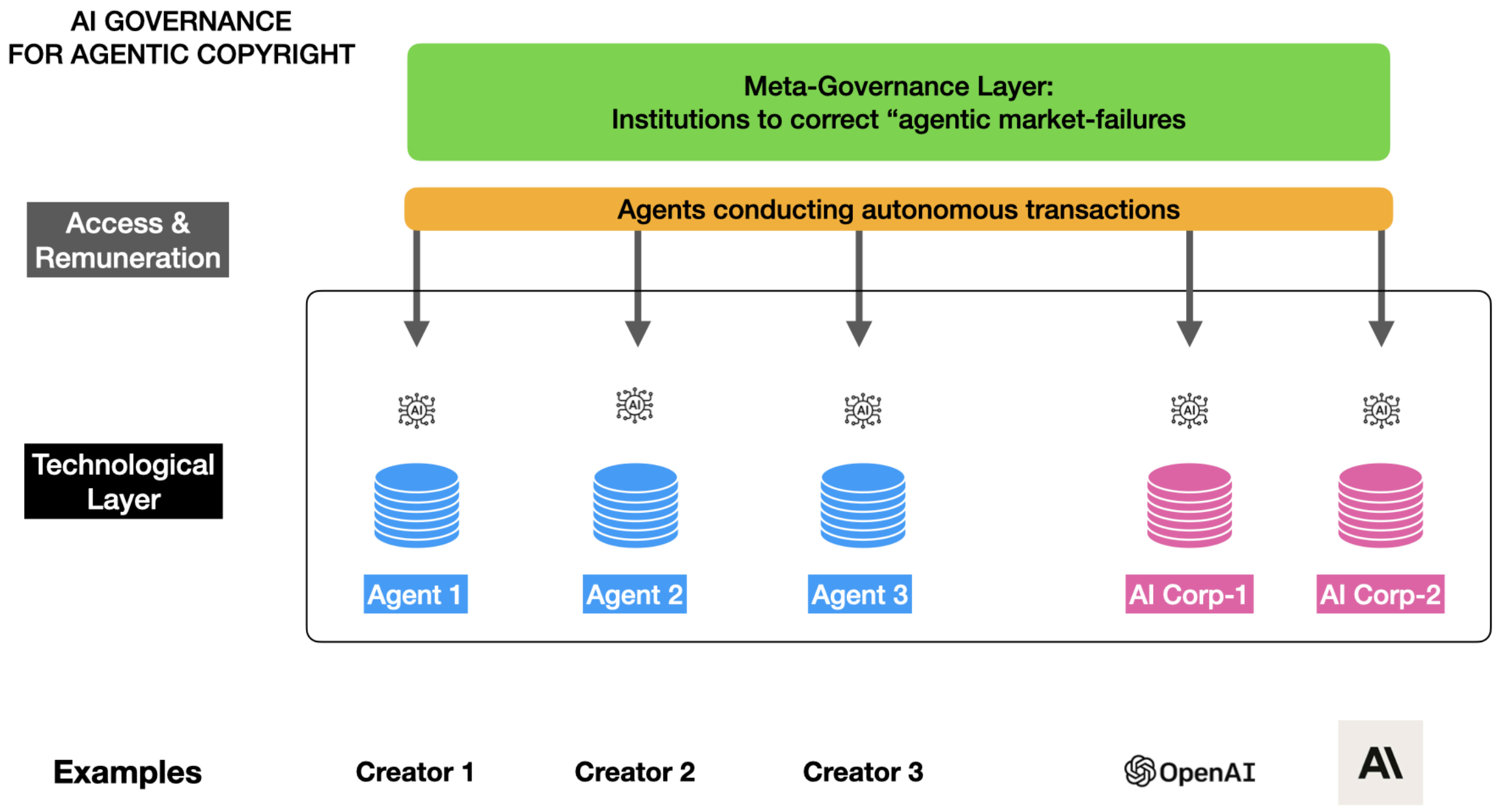

Each creator's "AI agent" negotiates with AI services (search bots, chat assistants, etc.) in microseconds, enforcing the creator's terms (like a smart copyright license). This AI-to-AI negotiation could allow individualized bargains at internet scale, moving beyond one-size-fits-all rules.

[109] Hammond, *supra* n. 82, p. 15.

### 3.3.1. Layer 1: Legal-Institutional Baseline

The first layer reflects a path-dependent perspective, recognizing that emerging agentic systems do not exist in a vacuum or outside of history. Rather, like any technology, agentic AI must be situated within existing laws and regulations. In creative industries, such existing legal frameworks point to copyright and intellectual property law, and, more broadly, the legal frameworks that form the foundation of the social contract, including constitutional law, property law, contract law, torts, administrative regulation, competition law, and consumer protection, among others that protect individual rights and sustain social cohesion.

Agentic copyright therefore operates within the current legal system. For example, even if AI agents handle licensing transactions, the legal validity of those agreements and their consequences must be assessed under existing statutes and case law. To the extent that new forms of agent-mediated interaction raise novel issues, the application of existing legal doctrines may require adaptation. Yet, the mere emergence of these new modes of agentic dealing, however, does not render them *per se* "extra-legal" or exempt from established legal principles. In this sense, Layer 1 is concerned with preserving the rule of law and the baseline social contract even as technological capabilities evolve.

### 3.3.2. Layer 2: Technical-Operational Layer

The middle layer concerns the governance of interactions among AI agents. It encompasses technical and organizational measures related to the design, deployment, and ongoing maintenance of AI agents and multi-agent systems. This layer includes standard protocols that enable agents to interact, negotiate, and transact, as well as mechanisms for identifying and mitigating technical risks such as security vulnerabilities, cascading failures, and adversarial behavior.[110] For developers, this layer will be the most familiar, as it focuses on system architecture, interoperability standards, audits, and safeguards embedded directly into the technology. Its function is to provide the "rails" on which AI agents operate, ensuring robustness and enabling the detection and remediation of technical failures. The design of this layer should involve not only developers, but also regulators, users, and broader societal stakeholders, to help ensure that agentic systems advance socially beneficial objectives.

This raises the deepest normative questions in our proposal, namely how to manage bargaining asymmetries between AI models. Clearly, this requires some procedural safeguards for agentic bargaining. It is not our argument that automation produces fairness, but rather that agentic copyright must be conditioned on a minimum bargaining architecture that includes inter alia baseline disclosure rules, auditable logs of negotiation, bounded offer-explanation duties, access to benchmark licensing terms and default constraints such as MFN-style protections or regulator approved safe-harbor protocols. In this way a creator-side fiduciary or aligned bargaining agent might be envisaged whose optimization is not the "raw" completion of a deal at any costs but human creator welfare under some predefined framework of preferences. Such safeguards become important, not least because they distinguish our model from a naïve, "let the bots bargain it out" framework. This acknowledges that the core move of our model is to focus on establishing a fair bargaining environment rather than the conclusion of deals per se.

---

110 Anthropic PBC, Introducing the Model Context Protocol (Nov. 25, 2024), https://www.anthropic.com/news/model-context-protocol.

### 3.3.3. Layer 3: Meta-Governance of Agentic Markets

The third layer aims to address risks arising from identified agentic market failures (miscoordination, conflict, and collusion). It consists of institutions and mechanisms designed to detect, correct, and prevent systemic failures in the agentic market as a whole. Such institutions may perform algorithmic auditing functions, for example by requiring major AI agent operators to submit pricing or negotiation algorithms for review in order to assess whether anticompetitive practices, such as price fixing, are embedded in their design.

This third "meta-governance" layer is arguably the most important, as it provides a framework for ensuring that multi-agent interactions remain aligned with underlying social values, including fairness, accountability, competition, the elimination of bias in automated decision-making, and respect for creative endeavor. Legal rules alone (Layer 1) are often too slow to keep pace with rapidly evolving agentic markets, while purely technical solutions (Layer 2) risk encoding narrow objectives without adequate attention to emergent ethical and institutional concerns. The meta-governance layer is therefore dedicated to remedying agentic market failures and serves as a bridge between law and code. It provides adaptive oversight, institutional memory, and value alignment in environments where autonomous AI agents increasingly transact, negotiate, and bargain on behalf of humans.

In the context of agentic copyright, such a governance layer is essential. As AI agents negotiate licenses, allocate remuneration, and manage access to creative works at scale, the primary risk is systemic distortion of creative markets. Addressing that risk goes beyond compliance with existing law (e.g., transparency and reporting obligations imposed by AI regulations). Meta-governance layer calls for collective governance mechanisms capable of observing emergent patterns, correcting undesirable equilibria, and ensuring that the benefits of agentic systems are distributed in a manner consistent with the purposes of copyright itself.

We situate this three-tier *Supervised Agentic Governance* framework within a broader social contract required for the AI age.[111] Rather than viewing AI as an existential threat to creative labor that must be restrained through blunt force,[112] we argue that the proposed framework can function as an enabler of new forms of value exchange in creative and data markets. In an agentic copyright environment, information exchange would increasingly occur through interactions *with, between, and via* AI agents, enabling creators to exercise greater dominion and control over their works than has previously been possible. Importantly, agentic copyright could help ensure that credit and economic value flow to their rightful owners, rather than to undeserving intermediaries. The aim, in other words, is to explore the possibility of a win–win outcome: AI systems continue to advance and deliver socially valuable services, while human creators are compensated and incentivized to produce high-quality content that improves those systems. This vision emphasizes mutual reinforcement between AI and human creativity, rather than a zero-sum tradeoff.

[111] Jurcys et al, *supra* n. 67.

# 4. Predicted Criticisms

Any plausible account of agentic copyright and supervised agentic governance must begin by taking its most serious objections on their own terms. The proposed framework will predictably be met with three concerns: that the marginal value of individual works in large-scale AI training is often negligible, that copyright intermediaries continue to perform functions that low-friction bargaining cannot replace, and that a stronger emphasis on authorization and control may come at the expense of innovation and social welfare.

These objections are substantial, but they do not foreclose the project. On the contrary, they help identify the settings in which agentic copyright is least persuasive, the contexts in which it may have genuine practical force, and the normative tradeoffs that any credible copyright framework for the AI era must confront. The aim of this chapter 4 is therefore to explore and specify the limits of agentic copyright, its institutional complements, and its potential contribution to a more adaptive law of creative exchange.

## *4.1. Marginal Value Problem and Emerging New Data Markets*

It is well established that, in the context of large-scale foundation-model training, the marginal value of any single work within a massive training corpus is often close to zero.[113] That basic economic reality has shaped much of the current debate over copyright and AI. It also explains why simple "work-by-work" licensing models may fail to scale in the AI training context.

---

[112] Some call to destroy LLMs because they were trained on copyrighted content, see e.g., NYT v OpenAI complaint, *supra* n. 18 (asking the court to order the destruction under 17 U.S.C. § 503(b) of all GPT or other LLM models and training sets that incorporate Times Works).

[113] See e.g., Matthew Sag, The False Hope of Content Licensing at Internet Scale, ProMarket (Nov. 19, 2025), https://www.promarket.org/2025/11/19/the-false-hope-of-content-licensing-at-internet-scale/; Sheng Zha et al., The Economics of AI Training Data: A Research Agenda, arXiv:2510.24990, at 3 (2025).

Our claim is not that agent-to-agent bargaining will routinely generate meaningful per-work compensation for every copyrighted work incorporated into model training. Rather, we offer the more modest claim that agentic systems may reduce the costs of search, verification, preference expression, monitoring, and compliance across a broader set of licensing environments. This is especially the case where value turns not on any single work in isolation, but on aggregation, curation, provenance, exclusivity, recency, modality-specific quality, or downstream use limitations.

That distinction matters because foundation-model training differs in important respects from ordinary bilateral licensing of individual works. In conventional copyright markets, parties often negotiate over a discrete asset or a clearly bounded bundle of rights. In the AI-training and related contexts, by contrast, the relevant economic unit is frequently not the individual work at all, but the corpus, subset, category, or, more importantly, the entire access regime. Once the inquiry into agent-to-agent bargaining is framed at that level, the limits of atomized bargaining become more apparent.

Emerging data markets already indicate that we are entering a post-training AI environment in which agent-to-agent bargaining becomes increasingly salient. AI systems and autonomous data-scraping agents are creating genuine commercial demand for access-controlled data environments.[114] In this context, the central question is not the isolated value of any individual work, but one of scale: how AI agents may facilitate lawful access to well-structured pools of information, and how such agentic interactions should be organized, authenticated, and governed.

For that reason, the market for frontier model training on massive corpora should be understood as only one part of a broader and more differentiated data ecosystem. Agentic bargaining may be weakest where access depends on vast, diffuse, and substitutable datasets for which individual contributions are economically negligible. But the analysis changes in markets for domain-specific or premium datasets, where curation, provenance, legality, authenticity, or recency carry independent value. It changes again in downstream licensing environments, including fine-tuning, retrieval, branded or authenticated corpora, permissions (e.g., access and use of synthetic voices or images).[115] In these new data markets control over access, inputs and outputs becomes more granular and commercially salient.[116] In those settings, agentic systems may have greater practical force because they can help structure permissions, attach use conditions, verify provenance, and lower the frictions associated with repeated or dynamic licensing.

[114] X Corp. v. Bright Data Ltd., No. 23-cv-03698-WHA (N.D. Cal. May 9, 2024); Reddit, Inc. v. SerpApi LLC et al., No. 1:25-cv-08736 (S.D.N.Y. filed Oct. 22, 2025); and Amazon.com Services LLC v. Perplexity AI, Inc., No. 3:25-cv-09514-MMC (N.D. Cal. Mar. 9, 2026) (preliminary injunction order).
[115] See e.g., ElevenLabs, Deploy AI Agents in Minutes, Not Months, https://elevenlabs.io/agents (last visited Apr. 3, 2026).

## *4.2. Agentic Copyright Does Not Substitute Existing Intermediaries*

Agentic copyright requires more careful treatment of intermediation. The problem with traditional copyright intermediaries is not simply that they need to be dominant to reduce transaction costs and facilitate information exchange.[117] Publishers, labels, collecting societies, and other market actors also provide capital, absorb risk, coordinate marketing and distribution, aggregate audiences, and confer reputational validation. Those functions do not disappear merely because agent-to-agent negotiation becomes cheaper. Accordingly, the stronger argument is not that AI agents render traditional intermediaries obsolete, but that they may unbundle certain intermediary functions while leaving others intact. Even in a world of low-friction agentic exchange, creators may still rationally prefer intermediaries that finance discovery, bear market uncertainty, or amplify audience reach.

On that view, agentic copyright does not eliminate intermediation. By making some data access and licensing functions cheaper and more contestable, agentic systems may discipline the market by pressuring incumbents to justify their margins function by function, rather than relying on legacy control over access, administration, and market coordination. In other words, the agent-to-agent bargaining model is not presented here as a wholesale substitute for, or replacement of, existing intermediaries and prevailing IP licensing structures. Rather, it is conceived as a viable mechanism for a subset of (new) bargaining transactions that may arise in an AI-powered environment, in which autonomous agents search for, exchange, and negotiate over information capable of being deployed for a wide range of commercial purposes within an increasingly sophisticated digital infrastructure.

The growing ubiquity of AI agents across commercial settings gives rise to an obvious bootstrapping problem. Early adopters would incur meaningful opportunity costs were they to forgo established licensing arrangements before the market had matured sufficiently to render agentic licensing economically viable. For that reason, agentic copyright is best understood as capable of emerging in at least three forms: first, as a complement to existing publishers and other information intermediaries; second, as a parallel market channel for agent-to-agent exchanges of information; and only later, and in more limited sectors, as a potential substitute for legacy licensing arrangements.

---

[116] Jurcys et al., *supra* n. 70 (exploring various layers of licenses for access and use of personal data).

[117] Shinto Teramoto & Paulius Jurcys, Intermediaries, Trust and Efficiency of Communication: A Social Network Perspective, in Networked Governance, Transnational Business and the Law 99, 99 (Mark Fenwick et al. eds., 2014).

Recognizing the rapid development of the agentic ecosystem also helps illuminate the network effects likely to shape its adoption. The most plausible early use cases are therefore unlikely to involve the immediate displacement of major publishers, labels, or other incumbent intermediaries. More realistically, the model is likely to gain traction first in settings such as fragmented digital uses or machine-readable permissions markets, where transactional complexity is high, rights are dispersed, and traditional licensing mechanisms are non-existent or comparatively inefficient.

## *4.3. Fairness v Welfare*

One way to understand the normative stakes of agentic copyright is through the familiar tension between fairness and welfare.[118] A fairness-oriented account begins from an intuitively powerful premise suggesting that if copyrighted works are used in AI systems without meaningful authorization, attribution, or compensation, the law should correct that perceived imbalance.[119] On that view, the primary objective is to preserve control, vindicate entitlement, and ensure that creators are not treated merely as inputs into a larger technological system. But that perspective, while normatively appealing, does not exhaust the inquiry. As the literature on fairness and welfare reminds us, legal rules must also be assessed in terms of their broader social consequences, including their effects on innovation, access, and the production of new forms of value.

That tension becomes especially acute in the age of agentic AI. If the law becomes singularly preoccupied with fairness in the narrow sense of perfect ex ante permission or individualized compensation for every use, it risks constructing a rights architecture so rigid that it suppresses socially valuable innovation. In the context of AI, the welfare gains may be immense: lower-cost knowledge production, new creative tools, greater access to information, and the emergence of entirely new markets for expression and exchange. From that perspective, a copyright system designed solely to insulate every right holder from any unconsented use may impose costs that are far greater than the harm it seeks to prevent. The point is not that fairness is irrelevant, but that an excessive commitment to fairness, abstracted from consequences, may produce a chilling effect that undermines the larger public purposes copyright is meant to serve.

[118] Kaplow & Shavell, Fairness versus Welfare (Harvard Univ. Press 2002).

[119] See e.g., Mark A. Lemley & Bryan Casey, Fair Learning, 133 Tex. L. Rev. 445, 482 (2024); Pamela Samuelson, Justifications for Fair Uses, 122 Mich. L. Rev. 299, 326 (2023); Molly Shaffer Van Houweling, The Freedom to Extract in Copyright Law, 103 N.C. L. Rev. 445 (2025); also Matthew Sag, Fairness and Fair Use in Generative AI, 92 Fordham L. Rev. 1887, 1900 (2023).

Agentic copyright should therefore move away from efforts to maximize fairness at all costs; it should attempt to mediate between fairness and welfare under new technological and socio-economic conditions. Its promise lies precisely in the possibility that computational agents can reduce transactional frictions, improve preference matching, and create more calibrated forms of licensing and access, without requiring the legal system to choose between absolute control and unrestricted appropriation. Properly designed, agentic systems may make copyright markets more responsive, granular, and administrable, while preserving enough flexibility to sustain experimentation and downstream innovation. The deeper claim, then, is that the future of copyright in an agentic environment should not be governed by fairness alone. Kaplow & Shavell's seminal research[120] suggests that copyright law must also account for welfare, including the incalculable public benefits that may flow from permitting socially productive uses of protected works under conditions that remain bounded, transparent, and institutionally accountable.

How would this "agentic copyright" framework apply to a more niche case of schools or libraries that cover the costs for making the copyright-protected content accessible to a large pool of users? Would these schools/libraries/companies now have to give each student/user a budget per year that their agent could use to negotiate prices? This use case is instructive because it exposes that not all licensing models revolve around bargaining between individual users and IP rights-holders. Some domains depend on pooled access, blanket licensing, and institutional budgeting. A virtue of the proposed "agentic copyright" framework is that it can operate at multiple levels of aggregation. In consumer or creator-to-platform markets, the licensing may be highly individualized. But in educational, library, and enterprise settings, agentic systems could just as easily be used by institutions to negotiate blanket or portfolio licenses on behalf of users. So, the answer is not that every student gets a micro-budget and an autonomous bargaining bot. Rather, schools, libraries, and companies could deploy institutional agents to procure collective access rights, allocate usage internally, and monitor compliance. That preserves the economic logic of blanket access while still benefiting from machine-readable negotiation and auditability.

# 5. Paths Forward

As generative AI has become the common denominator of our information society, the tension between innovation and creativity cannot be ignored. The early approaches to compensation, such as a blanket fair use on one side, or attempting to sue and legislate AI into submission on the other are unlikely to yield a satisfying, long-term solution. The U.S., with its fair use doctrine, might well conclude that using copyrighted data to develop AI is (in certain cases) lawful, especially given the transformative potential of AI technology and historical analogies like search engine caching and Google Books.[121]

[120] *Supra* n. 118.

Such a result would, no doubt, thrill AI developers and those who see AI progress as a national security priority, but it could also send a chilling message to creators who fear that their works are being taken without asking. Conversely, a regime of statutory licensing or an expansive new exclusive right over data mining could hand content owners (especially large ones) a tool to tax and control AI development at every turn.[122] That might funnel some money to publishers in the short term, but it could also entrench incumbents, exclude smaller players (both small creators and small AI start-ups), and slow technological advancement.

One of the possible solutions to escape this apparent deadlock lies in rethinking how AI can be harnessed to address complex copyright dilemmas that are emblematic of the age of AI. The analysis in this paper aims to point toward a Coasean path forward: when transaction costs fall sufficiently, AI-agent mediated private ordering can produce more efficient and equitable outcomes that blunt, all-or-nothing legal rules often miss. AI is frequently framed as the source of copyright's disruption, but it may also supply the tools for its repair by enabling frictionless, fine-grained bargaining over attribution, access, and compensation. Empowering creators and other market players with AI agents offer a way to level the playing field, allowing individual authors to participate meaningfully in markets historically dominated by large intermediaries, while preserving flexibility for AI developers to license what they value at prices the market deems fair. Rather than choosing between blanket fair use that alienates creators and heavy-handed licensing regimes that risk entrenching incumbents, agentic copyright proposes a middle course in which technology mediates rights in practice, not just in theory.

The proposed supervised agentic governance approach seems to offer a fresh perspective to copyright's incentive structure without abandoning it, aligning with a pragmatic vision in which law sets the guardrails and markets do the work. If successful, agentic copyright could help transform today's impasse into a cooperative equilibrium, where human creativity and AI innovation reinforce one another in service of progress, cultural development, and the public good.

---

[121] See e.g., Bartz v. Anthropic PBC, No. 3:24-cv-02200-WHA (N.D. Cal. June 24, 2025); Kadrey v. Meta Platforms, Inc., No. 3:23-cv-03417-VC, 2025 WL 3288499, at 8 (N.D. Cal. June 25, 2025). See also Edward Lee, Fair Use and the Origin of AI Training, 63 Hous. L. Rev. (forthcoming 2026), available at https://papers.ssrn.com/sol3/papers.cfm?abstract_id=5253011.

[122] Martin Senftleben, Win-Win: How to Remove Copyright Obstacles to AI Training While Ensuring Author Remuneration (and Why the AI Act Fails to Do the Magic), 100 Chi.-Kent L. Rev. 7 (2025); Senftleben, Generative AI and Author Remuneration, 54 IIC 1535 (2023); Szkalej & Senftleben, Generative AI and Creative Commons Licences: The Application of Share Alike Obligations to Trained Models, Curated Datasets and AI Output, 15 JIPITEC 313 (2024).